%% file: duojepa_paper_v2.tex
\documentclass[11pt]{article}
\usepackage[a4paper,margin=2.5cm]{geometry}
\usepackage[T1]{fontenc}
\usepackage[utf8]{inputenc}
\usepackage{lmodern}
\usepackage{amsmath,amssymb}
\usepackage{bm}
\usepackage{graphicx}
\usepackage{booktabs}
\usepackage{multirow}
\usepackage{array}
\usepackage{adjustbox}
\usepackage{xcolor}
\usepackage{tikz}
\usepackage{placeins}
\usepackage{caption}
\usepackage[numbers,sort&compress]{natbib}
\usepackage[colorlinks=true,linkcolor=blue!50!black,citecolor=blue!50!black,urlcolor=blue!50!black]{hyperref}

\newcommand{\R}{\mathbb{R}}
\newcommand{\cls}{\operatorname{cls}}
\newcommand{\LN}{\operatorname{LN}}
\newcommand{\Lc}{\mathcal{L}}
\newcommand{\om}{\bar{m}}
\newcommand{\rr}[2]{#1\,/\,#2}
\newcommand{\B}[1]{\textbf{#1}}
\DeclareMathOperator{\clip}{clip}
\DeclareMathOperator{\SIGReg}{SIGReg}

\title{DuoJEPA: Cross-Modal Translation Between\\Unaligned Self-Supervised Embedding Spaces}
\author{Mahmut S. Gokmen\textsuperscript{1}, Mitchell A. Klusty\textsuperscript{1}, Evan W. Damron\textsuperscript{2},
  Caroline N. Leach\textsuperscript{2},\\
  Samuel E. Armstrong\textsuperscript{1}, V.~K. Cody Bumgardner\textsuperscript{1}\\[1.5ex]
  {\small \textsuperscript{1}Kentucky Center for Digital Innovation, University of Louisville, Louisville, KY, United States}\\
  {\small \textsuperscript{2}University of Kentucky, Lexington, Office of Research, College of Medicine, United States}}
\date{}

\begin{document}
\maketitle

\begin{abstract}
Learning from paired modalities is commonly equated with aligning them: two encoders are trained until the embeddings of a pair
coincide in a shared space. We study whether cross-modal retrieval and classification can instead be obtained while the two
encoders' own embedding spaces remain directly unaligned. DuoJEPA trains one encoder per modality with the self-supervised
LeJEPA objective and appends to each encoder a single transformer layer that translates its embedding into the other encoder's
space, supervised by a contrastive objective computed entirely within that target space. No loss term compares the two own
spaces directly, although the encoders are optimised jointly through the translation objective. In our experiments, the
translator's residual connections determine where the cross-modal pressure is absorbed: with a residual translator the two
encoders become directly aligned, whereas with a skipless (residual-free) translator a paired-cosine test detects no alignment
in the final models, and at most a weak one in the ablations, at comparable cross-modal performance. The learned class-token
translation is well approximated by a single scaled rotation of the source space. We evaluate DuoJEPA on two modality pairs that
differ in nearly every respect. On chest radiographs and reports from CheXpert Plus, it attains a zero-shot AUC of 86.6 and a
cross-modal retrieval score of 78.5, compared with 60.0 and 64.2 for our reimplementation of LeVLJEPA with the same encoders and
data. On Sentinel-2 optical and Sentinel-1 radar imagery, the frozen encoders fall within the range of published Sentinel
foundation models under our reproduction of the Copernicus-Bench linear-probe protocol, the translator retrieves the paired
image of the other sensor among 4,000 candidates in 77--79\% of cases, and a patch-level extension transfers the within-image
similarity structure of the scene.
\end{abstract}

\section{Introduction}
\label{sec:intro}

Many imaging problems provide a second, complementary observation of the same object. Radar and optical satellites image the
same location within days of each other, and every chest radiograph is accompanied by the report in which a radiologist
describes it. The second observation contains information that the first lacks: synthetic aperture radar is insensitive to cloud
cover and illumination, and a report names findings that the image only depicts. Representation learning maps each input to a
vector, its \emph{embedding}, through a neural network, the \emph{encoder}. The prevailing approach to paired observations is
contrastive alignment, in which two encoders are trained such that the embeddings of a pair coincide and those of different
pairs are separated, until both modalities occupy a shared space
\cite{radford2021clip,zhai2023siglip,huang2021gloria,tiu2022chexzero,fuller2023croma}. The approach is simple and effective,
since cross-modal retrieval and the recognition of a textually described finding in an image both reduce to a nearest-neighbour
search in the shared space.

Alignment nevertheless imposes a constraint. Each encoder must express its modality in coordinates shared with the other.
Information carried by only one modality competes with the shared component for the same coordinates, and even a successfully
trained contrastive model leaves the two modalities in distinct regions of the shared space \cite{liang2022mindthegap}. Methods
that aim to preserve modality-specific information therefore reserve part of the embedding for it \cite{wang2024decur}. Recent
non-contrastive methods dispense with negative pairs but retain a shared prediction target; in LeVLJEPA, for instance, image and
text embeddings predict each other through separate predictor networks \cite{levljepa}. These observations motivate the
question addressed in this work: whether paired data can support cross-modal retrieval and classification while the two
encoders' own embedding spaces remain directly unaligned.

Research on the geometry of learned representations suggests that this may be possible. Networks trained independently on related
data tend to agree on which inputs are similar, even when their coordinates are unrelated
\cite{kornblith2019cka,huh2024platonic,maniparambil2024vision}. Translation between embedding spaces rests on this regularity.
The word embeddings of two languages are related to a good approximation by an orthogonal map
\cite{mikolov2013exploiting,xing2015normalized,conneau2018word}, the embedding spaces of independently trained vision and language
models can be translated into one another by simple transformations estimated from a small number of paired points
\cite{moschella2023relative,maiorca2023latent,norelli2023asif}, and recent methods relate embedding spaces without any paired data
\cite{jha2025vec2vec,schnaus2025blind,schnaus2026rosetta}. In all of these cases the encoders are trained first and the mapping is
estimated afterwards. We adopt as a working hypothesis that two encoders trained on the same samples can arrange them into similar
neighbourhoods within differently oriented frames, in which case pairing can be used to learn a mapping between the two spaces
rather than to merge them. If this holds, the mapping need not be estimated after training: it can be learned together with the
encoders.

We introduce \emph{DuoJEPA}, a paired pre-training method built on this hypothesis. Each modality is assigned its own encoder,
trained with LeJEPA \cite{balestriero2025lejepa}, a self-supervised objective that requires neither labels, negative pairs nor a
teacher network. Each encoder is extended by one additional transformer layer, the \emph{translator}, which is trained such that
the translated embedding of a sample identifies the corresponding embedding of the other modality among those of other samples.
This objective is evaluated entirely within the space of the other encoder, so that no loss term compares the two own spaces
directly. The encoders are nonetheless trained jointly: the gradient of the translation objective also reaches the target
encoder, so the two spaces are not learned independently in the optimisation sense.

\begin{figure}[!b]
  \centering
  \includegraphics[width=\linewidth]{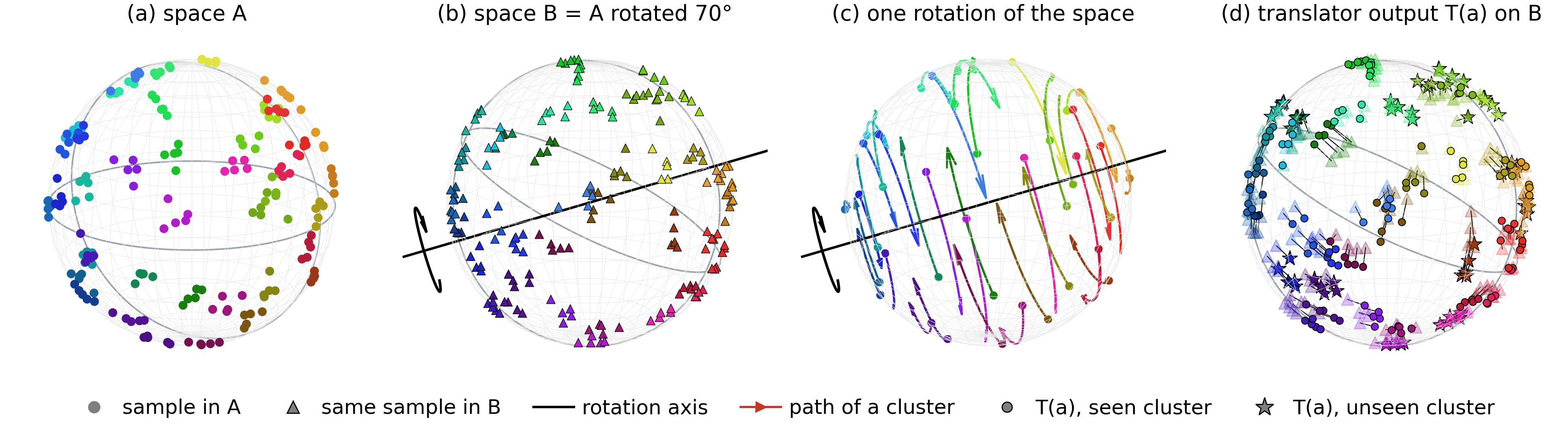}
  \caption{A simplified skipless translator trained with an InfoNCE objective on 27 of 40 synthetic clusters recovers a rotation
  planted between two three-dimensional spaces (a, b), which acts as one turn of the whole space (c), and maps the 13 clusters it
  never saw onto their targets (d); the setup is given in Appendix~\ref{app:toy}.}
  \label{fig:toy}
\end{figure}

The residual connections of the translator turn out to matter. A standard transformer layer adds its output to its input, so
that a translator built from such a layer is initialised close to the identity map. In our experiments, a residual translator led
the two encoders to become directly aligned, which suggests that the translation loss was reduced by moving the encoders into
common coordinates rather than only by learning a mapping. We therefore remove both residual connections from the translator
layer, using the residual-free (\emph{skipless}) transformer layer and initialisation of \citet{ji2025skipless}. With this layer,
any correspondence between the two spaces has to be computed by the translator. In the final models of both applications studied
here, the skipless translator kept the own spaces unaligned under a paired-cosine test at a cross-modal performance comparable to
that of the residual translator, and the learned translation of the class token is well approximated by a single scaled rotation of the
source space (Figure~\ref{fig:toy} illustrates the idea on synthetic data); this relation was much weaker in a control trained
without the translation objective. In concurrent work, \citet{balestriero2026platonic} prove that JEPAs trained separately on two
modalities learn the same representation up to a rotation when their positive pairs are two observations of the same scene; our
experiments arrive at a rotational relation from the empirical side, between encoders trained jointly through the translation
objective.

We evaluate DuoJEPA on two types of paired data that differ in nearly every respect: chest radiographs with their reports
\cite{chambon2024chexpertplus}, and Sentinel-2 optical with Sentinel-1 radar images of the same locations \cite{wang2023ssl4eo}.
On the former, DuoJEPA attains a zero-shot AUC of 86.6 and a cross-modal retrieval score of 78.5, compared with 60.0 and 64.2 for
our reimplementation of LeVLJEPA with the same encoder bodies and data, while the paired-cosine test detects no alignment between
its own spaces. On the latter, evaluated on BigEarthNet \cite{clasen2024reben} under our reproduction of the Copernicus-Bench
linear-probe protocol \cite{wang2025copernicusfm}, the frozen encoders reach 76.7 (radar) and 81.7 (optical) micro mAP, within the
range of published Sentinel foundation models. The translation allows a classifier trained on one sensor to process the other and
transfers the within-image similarity structure of the scene at the level of patches.

The contributions of this work are as follows.
\begin{itemize}
  \item We introduce DuoJEPA, a paired pre-training method that retains one self-supervised embedding space per modality and
        learns a translation between them within each encoder, jointly with the encoders. A single recipe serves image-text and
        image-image pairs, with a patch-level extension for images acquired on a common pixel grid.
  \item We show empirically that the residual connections of the translator determine whether the two encoders become directly
        aligned, and that a single skipless translator layer, using the residual-free design of \citet{ji2025skipless}, keeps
        them unaligned under a paired-cosine test in the final models of both applications, with at most a weak alignment in
        one ablation.
  \item We characterise the learned class-token translation as approximately a single scaled rotation of the source space, which
        explains most, but not all, of the translated embedding, and find that this relation between the own spaces is much
        weaker in a control trained without the translation objective.
  \item We evaluate the method on chest radiographs with reports and on optical and radar satellite imagery, and ablate each
        component of the recipe, with the limitations of single-run ablations stated explicitly.
\end{itemize}

\section{Background}
\label{sec:background}

This section introduces, for readers outside representation learning, the concepts on which the remainder of the paper relies.
Readers familiar with vision transformers and contrastive learning may proceed to Section~\ref{sec:related}.

\paragraph{Encoders and embedding spaces.}
An encoder is a neural network $f$ that maps an input, such as an image or a text, to a vector $f(x) \in \R^{d}$, its embedding;
throughout this paper $d = 768$. The set of embeddings an encoder can produce constitutes its \emph{embedding space}, in which
similarity is measured by the cosine of the angle between two vectors. A useful encoder places inputs of similar content close to
each other. This property is commonly assessed with a \emph{linear probe}: the encoder is kept fixed and a single linear classifier
is trained on its embeddings to predict labels. High probe performance indicates that the information required for the task is
present in the embedding and linearly accessible.

\paragraph{Vision and text transformers.}
All encoders in this work are transformers \cite{vaswani2017attention}. A vision transformer (ViT) \cite{dosovitskiy2021vit}
partitions an image into square patches, here of $14 \times 14$ pixels, so that a $224 \times 224$ image yields a grid of
$16 \times 16$ patches. Each patch is projected to a vector, a \emph{token}, and an additional learnable token, the
\emph{class token} (CLS), is prepended to the sequence. A stack of layers then exchanges information between tokens through
self-attention. After the final layer the class token summarises the input, while each patch token describes its region of the
image in the context of the whole. A text transformer such as BERT \cite{devlin2019bert} operates in the same manner on sub-word
tokens. Each layer adds its output to its input, $x \mapsto x + F(x)$; this \emph{residual connection} \cite{he2016resnet} allows
information to pass through a layer unchanged, a property that is central to the design of the translator
(Section~\ref{sec:skipless}).

\paragraph{Self-supervised learning.}
Self-supervised learning trains an encoder on unlabelled data. A common strategy generates several randomly perturbed
\emph{views} of the same input, for instance crops at different scales or photometric changes, and requires the encoder to assign
similar embeddings to all views of an input \cite{chen2020simclr,caron2021dino}. This requirement alone admits a trivial solution,
an encoder that maps every input to the same vector, a failure mode known as \emph{collapse}; a milder form, \emph{dimensional
collapse}, concentrates the embeddings on a few directions. LeJEPA \cite{balestriero2025lejepa} prevents collapse with a
regulariser, SIGReg, that compares the embeddings of a batch, projected onto random directions, with a standard Gaussian
distribution. In LeJEPA, SIGReg acts on the output of a small projector network appended to the encoder during training, not on
the encoder's embedding itself.

\paragraph{Contrastive learning from paired data.}
For paired modalities, the standard objective is InfoNCE \cite{oord2018cpc,radford2021clip}. In a batch of $B$ pairs, the
embedding of each image is compared with the embeddings of all $B$ texts; the similarities, divided by a \emph{temperature} $T$,
are converted into probabilities by a softmax, and the loss maximises the probability assigned to the image's own text, the
remaining $B - 1$ texts acting as negatives. Minimising this loss in both directions draws the two embeddings of a pair together in
a shared space. We call two encoders \emph{directly aligned} when the embeddings of a pair are closer to each other than to the
embeddings of other samples without any mapping between the spaces. Alignment enables \emph{zero-shot classification}, in which an
image is classified without labelled examples by comparison with short text prompts such as ``pleural effusion'' and ``no pleural
effusion'', and \emph{cross-modal retrieval}, in which the report belonging to an image is identified among many candidates.

\paragraph{Orientation of embedding spaces.}
The LeJEPA objective constrains the distribution of the projected embeddings but not their orientation. Its invariance term is
exactly invariant to an orthogonal transformation of the projected space, and SIGReg, which compares the batch with an isotropic
Gaussian along random directions drawn afresh at every step, is invariant to such a transformation in expectation over the
directions. Moreover, because the projector receives the encoder's class token without a preceding normalisation, an orthogonal
transformation of that class token can be compensated exactly by the projector's first linear layer. At the interface between
encoder and projector, the objective therefore does not prefer an orientation. This does not imply that the encoder itself is
indifferent to orientation, since its internal normalisation layers and fixed coordinate structure are not rotation-equivariant,
and in DuoJEPA the translation objective couples the two encoders. The resulting hypothesis is that two encoders trained on the
same samples can settle in frames whose relative orientation is arbitrary, so that a translation between them may be close to an
orthogonal map. Concurrent work reaches a related conclusion by a different route: \citet{balestriero2026platonic} prove that
JEPAs trained separately on two modalities learn the same representation up to one rotation when the positive pairs of each
modality are two observations of the same scene, whereas for positives that are corruptions of a single observation, such as two
crops of one image, the misalignment is only bounded. Figure~\ref{fig:toy} illustrates this situation on synthetic data, and Section~\ref{sec:rotation} tests it on the
trained models; there, the relation is strong between the jointly trained encoders and much weaker in a control trained without
the translation objective, which suggests that the joint objective produces it rather than merely exposing it.

\section{Related Work}
\label{sec:related}

\paragraph{Self-supervised and joint-embedding learning.}
Self-distillation methods such as DINO, iBOT and DINOv2 learn image representations by matching a student network to a momentum
teacher across augmented views \cite{caron2021dino,zhou2022ibot,oquab2024dinov2,darcet2024registers}, whereas redundancy- and
variance-based objectives avoid the teacher by regularising the statistics of the embedding directly
\cite{zbontar2021barlow,bardes2022vicreg}. Joint-embedding predictive architectures (JEPA) predict the representation of one part
of the input from another in representation space rather than in pixel space \cite{assran2023ijepa,bardes2024vjepa}. LeJEPA
\cite{balestriero2025lejepa} reduces this family to two terms, an invariance term that draws the views of a sample together and
SIGReg, and requires neither negatives nor a teacher. DuoJEPA trains each of its encoders with LeJEPA; the translation is the only
term that couples them.

\paragraph{Learning from paired modalities.}
CLIP \cite{radford2021clip} aligns images and captions with a symmetric InfoNCE objective, and subsequent work varies the
encoders, the loss and the number of modalities \cite{zhai2022lit,zhai2023siglip,girdhar2023imagebind}, aligns tokens rather
than global embeddings \cite{yao2022filip}, or complements the contrastive objective with self-supervised and non-contrastive terms
\cite{mu2022slip,zhou2023nclip}. In medical imaging, radiographs are aligned with their reports at the global, local or disease
level \cite{zhang2022convirt,huang2021gloria,boecking2022biovil,wang2022mgca,wang2022medclip}, and CheXzero demonstrates that the
resulting space supports zero-shot classification comparable to radiologists for several findings \cite{tiu2022chexzero}. A
recent group of methods dispenses with negatives: NOVA \cite{kuhn2026nova} predicts the embedding of a frozen clinical text
encoder from radiograph views under SIGReg, VL-JEPA \cite{chen2025vljepa} predicts text embeddings from visual input, TC-JEPA
\cite{huang2026tcjepa} conditions an image JEPA on captions, and LeVLJEPA \cite{levljepa} trains both encoders end to end with
cross-modal predictors and stop-gradient targets and evaluates its visual backbone on dense prediction as well as on global tasks.
In Earth observation, Sentinel-1 and Sentinel-2 pairs are available at scale \cite{schmitt2019sen12ms,wang2023ssl4eo,clasen2024reben},
and radar and optical images are combined through contrastive alignment and fusion \cite{fuller2023croma}, through redundancy
reduction with common and modality-specific dimensions \cite{wang2024decur}, or within multi-sensor foundation models trained by
masked reconstruction, seasonal contrast or self-distillation
\cite{manas2021seco,cong2022satmae,reed2023scalemae,szwarcman2024prithvi,tseng2025galileo,xiong2024dofa,wang2024softcon,
danish2025terrafm,wang2025copernicusfm}; \citet{lane2025genealogy} survey this field, and JEPA-style prediction has been applied to
remote-sensing retrieval \cite{choudhury2025rejepa}. These methods produce a shared, fused or predicted space for the two
modalities, whereas DuoJEPA retains two own spaces and learns a mapping between them.

\paragraph{Translation between representations.}
Independently trained networks have long been observed to learn related representations. Model stitching connects two networks
through a learned linear layer \cite{lenc2015equivariance,bansal2021stitching}, representation similarity measures quantify the
structure shared by two networks \cite{kornblith2019cka}, and the platonic representation hypothesis posits that the
representations of different modalities converge as models grow \cite{huh2024platonic,maniparambil2024vision}.
In concurrent work, \citet{balestriero2026platonic} turn this hypothesis into a theorem for JEPAs: encoders trained separately on each modality learn
the same representation up to one rotation when their positive pairs are two observations of one scene, and this rotation is
identified by paired samples but not by the first two moments of unpaired data. A related line of work uses such shared structure
to translate between embedding spaces after training. In word translation, the embeddings of two languages are mapped onto each
other by an orthogonal transformation estimated from a seed dictionary or without parallel data
\cite{mikolov2013exploiting,xing2015normalized,smith2017offline,conneau2018word}. Relative representations
\cite{moschella2023relative}, latent space translation \cite{maiorca2023latent}, latent functional maps \cite{fumero2024latent},
ASIF \cite{norelli2023asif} and linear image-to-text maps \cite{merullo2023linearly} extend this idea to pre-trained vision and
language models, and recent methods dispense with paired data altogether, translating between text embedding spaces
\cite{jha2025vec2vec} or matching and aligning vision and language representations \cite{schnaus2025blind,schnaus2026rosetta}.
Translation has also been used as a training signal rather than as a post-hoc tool. MCTN learns a joint multimodal representation
by translating a source modality into a target modality and back \cite{pham2019found}, learned translators between latent spaces
appear in unpaired image translation \cite{sendik2020crossnet}, and in video-text retrieval LaT \cite{bai2022lat} replaces the
joint space by translations with a cycle consistency. DuoJEPA differs from these approaches in where the mapping is learned and in
what it keeps separate. Post-hoc maps, whether estimated from pairs or without them, and model stitching relate encoders that are
already trained and leave them unchanged; MCTN uses the translation to form a joint representation; predictor-based JEPA methods
learn a cross-modal predictor jointly with the encoders but towards a stop-gradient target; contrastive methods learn no mapping
and align the encoders instead. DuoJEPA learns a transformer translator jointly with both encoders, places the translation loss in the target space with
a trainable target, and uses a skipless translator to retain directly unaligned own spaces.

\paragraph{Residual-free transformers.}
Residual connections maintain the trainability of deep networks \cite{he2016resnet}. In their absence, self-attention loses rank
with depth \cite{dong2021attention} and the forward signal degenerates \cite{noci2022signal}; modified attention, normalisation
and initialisation schemes restore the trainability of transformers without shortcuts \cite{he2023deep,he2024simplifying}.
\citet{ji2025skipless} train residual-free vision transformers stably through an initialisation derived from a conditioning
analysis, without changing the architecture, and report richer hierarchical representations and improved dense prediction; in
subsequent work, \citet{ji2026skipdiffusion} remove the residual connections from the encoder of a diffusion transformer and
report improved representations for classification, segmentation, object discovery and semantic correspondence. DuoJEPA adopts
the skipless layer and initialisation of \citet{ji2025skipless}, but for a different purpose: as a single translator layer whose
lack of an identity path changes how the model accommodates the cross-modal translation objective.

\section{Paired Modalities and Datasets}
\label{sec:data}

\begin{figure}[!t]
  \centering
  \includegraphics[width=\linewidth]{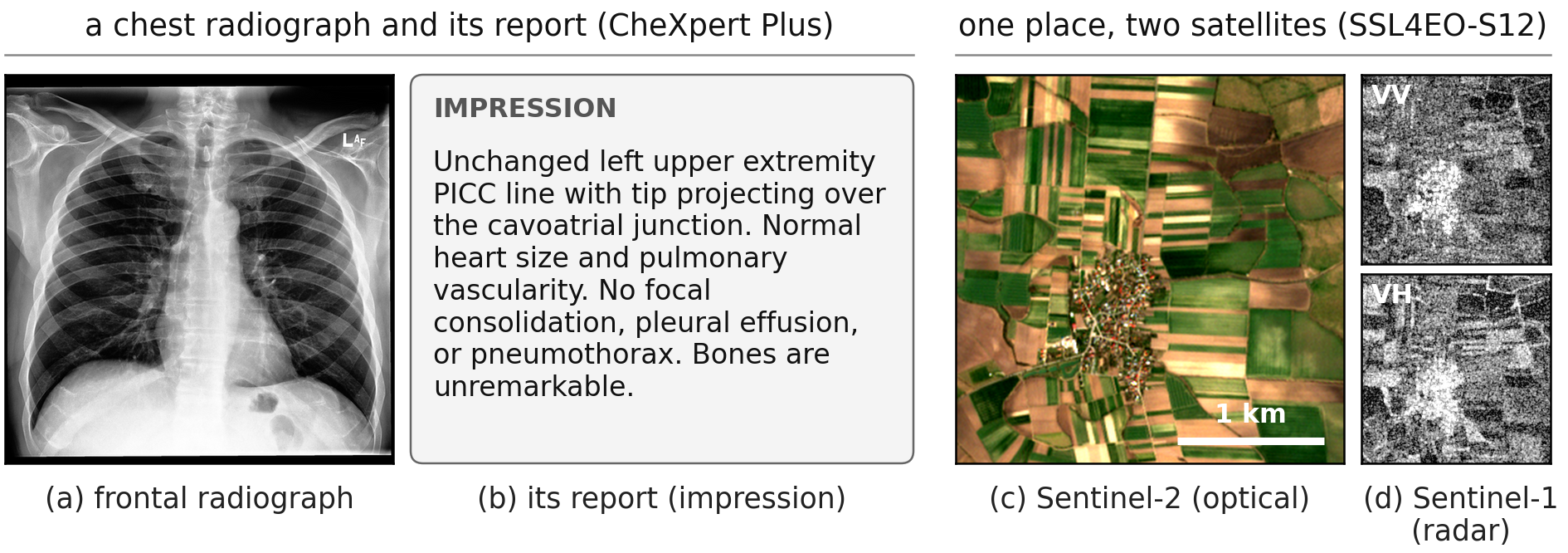}
  \caption{One pair of each kind: a frontal chest radiograph and the impression section of its report (a, b), and the same
  2.64\,km $\times$ 2.64\,km area seen by the optical Sentinel-2 in natural colour and by the radar Sentinel-1 in its two
  polarisations (c, d).}
  \label{fig:data}
\end{figure}

We study two pairs of modalities chosen to differ as much as possible (Figure~\ref{fig:data}, Table~\ref{tab:data}). In the
first, an image is paired with a text that describes it; in the second, two images of the same location are acquired by sensors
that measure different physical quantities.

\subsection{Chest radiographs and radiology reports}
\label{sec:data-cxr}

A chest radiograph is a projection X-ray image of the thorax. Because X-rays are attenuated more strongly by dense tissue than by
air, bones appear bright, the air-filled lungs dark, and the heart and vessels in intermediate grey levels
(Figure~\ref{fig:data}a). Much of the diagnostic content lies in subtle variations of texture and shape, such as an enlarged
cardiac silhouette or an opacity within the otherwise dark lung field. Each examination is documented in a free-text report. Its
\emph{findings} section lists the observations, and its \emph{impression} section summarises the conclusion in a few sentences,
frequently stating the absence as well as the presence of findings (Figure~\ref{fig:data}b).

We pre-train on CheXpert Plus \cite{chambon2024chexpertplus} (release file \texttt{df\_chexpert\_plus\_240401.csv}), which
provides the reports of the CheXpert collection \cite{irvin2019chexpert}: 223,228 chest radiographs paired with their reports,
from 187,711 studies of 64,725 patients. We keep the frontal images whose report contains an impression section, resized to 512
pixels on the short side and stored as 8-bit grey-level images, together with the impression truncated to 128 tokens. Evaluation
relies on four sets. The CheXpert test set \cite{irvin2019chexpert,tiu2022chexzero} comprises 500 radiographs labelled by
board-certified radiologists, on which, following common practice, we report five findings: atelectasis (partial collapse of lung
tissue), cardiomegaly (enlargement of the heart), consolidation (air spaces of the lung filled with fluid or cells), edema (fluid
in the lung) and pleural effusion (fluid surrounding the lung). For CheXpert 5x200 we draw, following the rule of GLoRIA
\cite{huang2021gloria}, 200 images from the CheXpert training set for each of these findings, each image being positive for that
finding only; since the draw of GLoRIA is not released, ours (seed 0) is a different sample of the same rule. Retrieval is
evaluated on the 202 image-report pairs of the CheXpert Plus validation split, and the alignment test of Section~\ref{sec:eval}
uses 1,859 image-report pairs from 645 patients held out from training. Patients are separated by construction: the 974 patients
of our 5x200 draw, the 645 hold-out patients and the validation patients are excluded from the pre-training set, which retains
183,140 images of 62,870 patients. The official test set is disjoint from the pre-training data at the patient level: its 500
patients carry identifiers (64,741 to 65,240) that occur nowhere in CheXpert Plus (1 to 64,740), whose de-identified patient
identifier corresponds one-to-one to these identifiers, and the CheXpert totals of 224,316 radiographs and 65,240 patients
\cite{irvin2019chexpert} equal the sums over its training, validation and test sets, so that the test patients are counted as
separate patients. Moreover, none of the 518 frontal test radiographs reappears among the 186,201 images we prepared: on
64$\times$64-pixel thumbnails, the highest correlation of a test radiograph with a prepared image is 0.964, whereas a validation
radiograph stored once as JPEG and once in our format correlates with itself at 0.998 or more (this comparison detects repeated
images, not repeated patients).

\subsection{Optical and radar satellite images}
\label{sec:data-eo}

The European Copernicus programme operates two complementary satellite constellations (Figure~\ref{fig:data}c, d).
\emph{Sentinel-2} carries a passive multispectral imager that records sunlight reflected by the surface in 13 spectral bands,
from the visible through the near infrared to the short-wave infrared, at 10, 20 or 60 metres per pixel. Its visible bands
approximate human vision, while its infrared bands distinguish, for example, vegetation from bare soil and water from land. As a
passive sensor, it depends on daylight and a cloud-free sky. \emph{Sentinel-1} carries an active synthetic aperture radar (SAR)
that emits microwave pulses in the C band (wavelength of approximately 5.6\,cm) and records the energy scattered back to the
satellite. Radar penetrates cloud cover and operates independently of illumination, but it measures a different physical property
of the surface: not reflectance, but roughness, geometry and moisture. Sentinel-1 records two polarisations, VV (vertical
transmission and reception) and VH (vertical transmission, horizontal reception), on a decibel scale, and its images are
affected by speckle, a granular interference pattern inherent to coherent imaging. The two sensors therefore describe the same
location in physically different terms, and the relation between them cannot be expressed as a pixel-wise transformation.

We pre-train on SSL4EO-S12 \cite{wang2023ssl4eo}, which samples 251,079 locations around the 10,000 most populated cities of the
world and images each of them in four seasons, yielding 1,004,316 Sentinel-1/Sentinel-2 pairs of 2.64\,km $\times$ 2.64\,km at 10
metres per pixel ($264 \times 264$ pixels). We use the 12 bands of the atmospherically corrected Sentinel-2 product (L2A; B1--B9,
B8A, B11, B12, reflectance divided by $10^4$) and the two Sentinel-1 polarisations (GRD, VV and VH in decibels). The two images of
a pair are matched by acquisition date (mean gap 6.25 days, maximum 29) and registered to a common pixel grid. Each band is
standardised with statistics estimated on 20,000 samples of the corpus. For evaluation we use BigEarthNet v2
\cite{clasen2024reben}, which contains 549,488 Sentinel-1/Sentinel-2 pairs of $120 \times 120$ pixels (1.2\,km $\times$ 1.2\,km) from
ten European countries, acquired between June 2017 and May 2018. Each pair is annotated with all land-cover classes it contains,
drawn from 19 classes derived from the CORINE Land Cover map, so that classification is a multi-label problem. We use the 480,038
pairs free of clouds, cloud shadows and seasonal snow, in their official split (237,871 training, 122,342 validation and 119,825
test pairs).

\begin{table}[!t]
\centering\small
\caption{The datasets used in this work.}
\label{tab:data}
\begin{adjustbox}{max width=\textwidth}
\begin{tabular}{lllll}
\toprule
dataset & modality pair & size used & sample & role \\
\midrule
CheXpert Plus \cite{chambon2024chexpertplus} & chest radiograph + report & 183,140 images, 62,870 patients & grey-level image + impression text & pre-training \\
CheXpert test, 5x200, validation, hold-out & radiograph (+ report) & 500 / 1,000 / 202 / 1,859 & grey-level image, finding labels & evaluation \\
SSL4EO-S12 \cite{wang2023ssl4eo} & Sentinel-2 + Sentinel-1 & 1,004,316 pairs, 251,079 locations & $264^2$ px at 10\,m, 12 + 2 channels & pre-training \\
BigEarthNet v2 \cite{clasen2024reben} & Sentinel-2 + Sentinel-1 & 480,038 clean pairs & $120^2$ px at 10\,m, 19 classes & evaluation \\
\bottomrule
\end{tabular}
\end{adjustbox}
\end{table}

\section{DuoJEPA}
\label{sec:method}

\begin{figure}[!t]
  \centering
  \includegraphics[width=\linewidth]{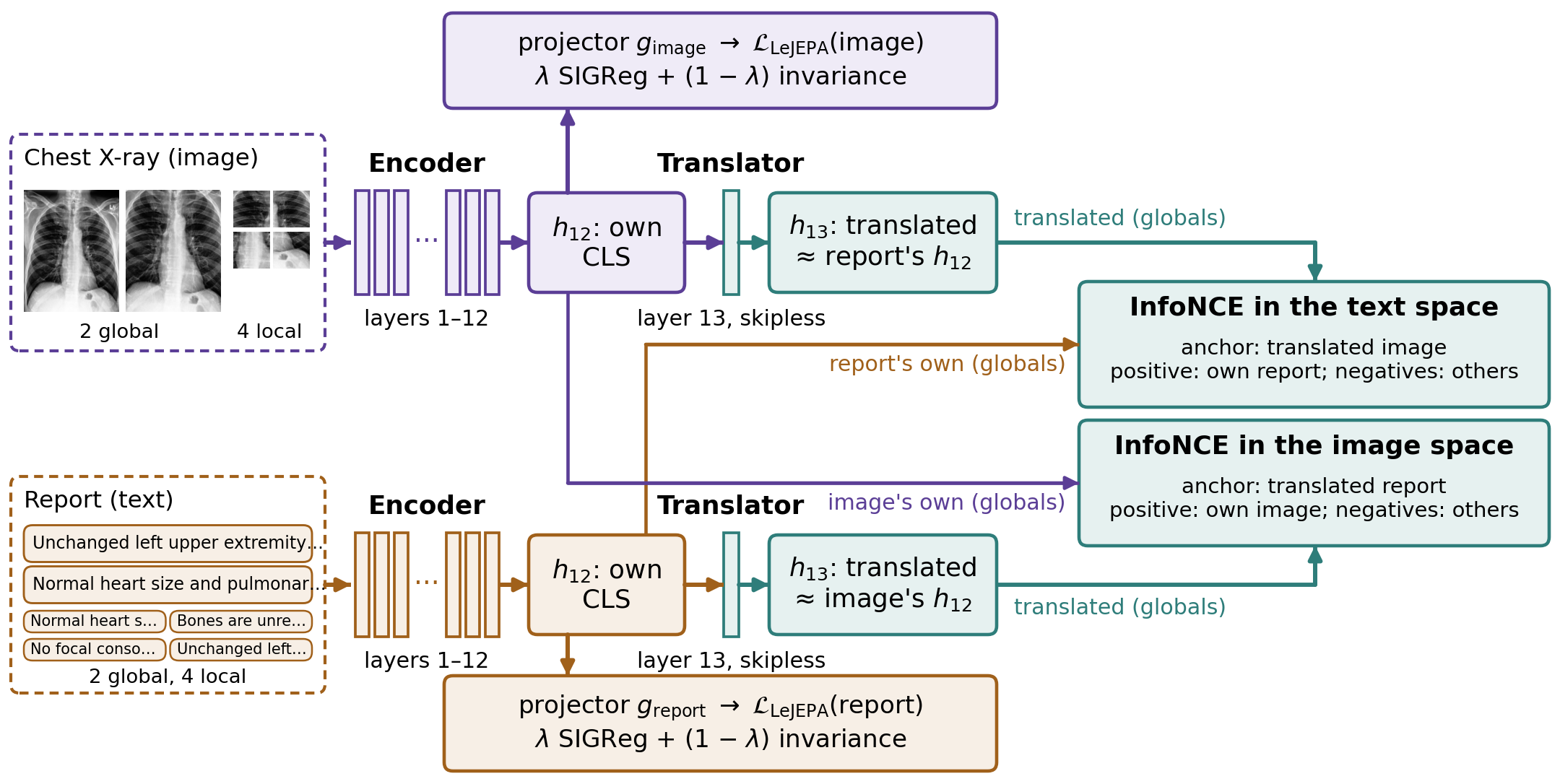}
  \caption{DuoJEPA on chest radiographs and their reports: each tower's own CLS ($h_{12}$) trains its LeJEPA loss, and its
  skipless translator layer ($h_{13}$) is trained by an InfoNCE in the other tower's space.}
  \label{fig:arch-vl}
\end{figure}

\begin{figure}[!t]
  \centering
  \includegraphics[width=\linewidth]{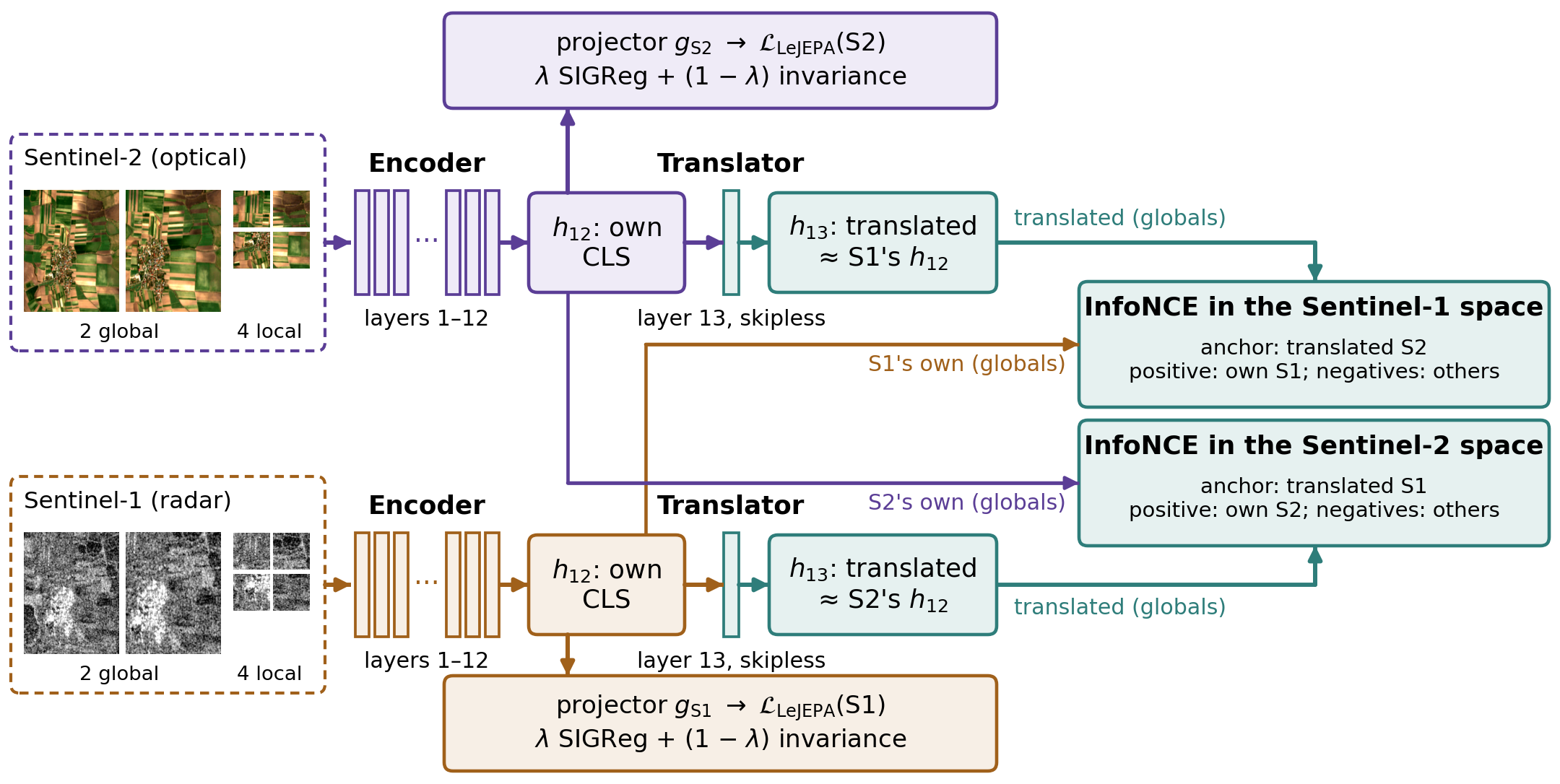}
  \caption{DuoJEPA on Sentinel-2 and Sentinel-1 images of the same place, with the same CLS route as Figure~\ref{fig:arch-vl}.}
  \label{fig:arch-vv}
\end{figure}

DuoJEPA consists of two encoders, one per modality, each followed by a translator layer that maps into the space of the other
encoder (Figures~\ref{fig:arch-vl} and~\ref{fig:arch-vv}). Each encoder is trained on its own modality, and the translators are
trained such that a translated sample can be identified among the own embeddings of the other modality. We first describe the
encoders and their readouts, then the translation objective and the translator layer, and finally the patch-level extension for
images acquired on a common pixel grid.

\subsection{Modality-specific encoders}
\label{sec:towers}

We consider pairs $(x^{a}_i, x^{b}_i)$ of the same sample in two modalities $a$ and $b$, and denote by $\om$ the modality other
than $m$. Each modality is processed by a \emph{tower} $f_m$, a transformer with $n = 12$ encoder layers followed by a single
translator layer. With $h^{m}_{\ell}(x)$ the token sequence after $\ell$ layers and $\cls(\cdot)$ its class token, a tower has
two readouts,
\begin{align}
  e_m(x) &= h^{m}_{12}(x), &
  \tau_m(x) &= h^{m}_{13}(x),
  \label{eq:readouts}
\end{align}
the \emph{own embedding} $e_m$, which lies in the own space of modality $m$ and serves all single-modality tasks, and the
\emph{translated embedding} $\tau_m$, which is trained to lie in the own space of $\om$. A cross-modal comparison places the
translated embedding of one modality next to the own embedding of the other, within the space of the latter. The translator
receives no conditioning signal and has no separate predictor network; its role as a translator is defined solely by the loss in
which its output appears. Because the translator is an ordinary layer of the tower, it is stored in the same checkpoint as the
encoder: each modality is saved as a single transformer of 13 layers, whose skipless computation is recorded in its configuration
(Appendix~\ref{app:impl}), and one forward pass yields both readouts. No separate prediction head or MLP is needed at inference; the
projector $g_m$ introduced below is used only during training and is not kept.

Each tower learns its modality with LeJEPA \cite{balestriero2025lejepa}. Augmentation produces, for each modality and sample,
$N = 2$ global and $K = 4$ local views $x^{m}_{i,v}$, with $v = 1, \dots, N$ global. For two images, the views of a pair are
cropped at the same location with the same flip, so that the two modalities of a view retain a common pixel grid; for a text, a
view is a subset of the sentences of the report. A projector $g_m$, a multilayer perceptron used only during training, maps the
own class token of every view, taken from $h^{m}_{12}$ without further normalisation, to $z^{m}_{i,v} \in \R^{p}$. The LeJEPA
loss of a tower combines SIGReg, evaluated on each view separately, with an invariance term that draws every view towards the mean
$\bar z^{m}_{i}$ of the global views of its sample:
\begin{align}
  \Lc^{m}_{\mathrm{LeJEPA}} &= \frac{\lambda_m}{N{+}K}\sum_{v=1}^{N+K} \SIGReg\big(Z^{m}_{v}\big)
  + \frac{1-\lambda_m}{(N{+}K)\,B\,p}\sum_{i=1}^{B}\sum_{v=1}^{N+K}\big\lVert z^{m}_{i,v} - \bar z^{m}_{i} \big\rVert^{2},
  \label{eq:lejepa} \\
  \bar z^{m}_{i} &= \frac{1}{N}\sum_{v=1}^{N} z^{m}_{i,v}, \nonumber \\
  \SIGReg(Z) &= \frac{1}{M}\sum_{j=1}^{M} B \int_{-5}^{5} \big\lvert \hat\varphi_j(s) - e^{-s^{2}/2} \big\rvert^{2} e^{-s^{2}/2}\, ds ,
  \qquad \hat\varphi_j(s) = \frac{1}{B}\sum_{i=1}^{B} e^{\,\mathrm{i}\, s\, \langle a_j, z_i\rangle} .
  \label{eq:sigreg}
\end{align}
Here $Z^{m}_{v}$ is the batch of projected embeddings of view $v$, $B$ the global batch size, and $a_1, \dots, a_M$ unit directions
drawn from an isotropic Gaussian afresh at every step; $\hat\varphi_j$ is the empirical characteristic function of the batch
projected onto $a_j$, averaged across devices, and is compared with that of a standard Gaussian by the Epps--Pulley statistic
\cite{epps1983test}. For a text tower, whose sentence-subset views can be empty, the invariance term is a mean over the valid views
only. The two terms act on the projected embeddings, not on the own embedding $e_m$ itself.

\subsection{Translation in the target embedding space}
\label{sec:cls}

The translation is learned separately in each of the two own spaces. In the space of $m$, the anchors are the translated
embeddings of the global views produced by the other tower, and the positive of an anchor is the centroid of the global views of
the same sample in $m$:
\begin{equation}
  q^{m}_{i,v} = u\big(\cls\, \tau_{\om}(x^{\om}_{i,v})\big), \qquad
  c^{m}_i = \frac{\sum_{v=1}^{N} u\big(\cls\, e_m(x^{m}_{i,v})\big)}{\big\lVert \sum_{v=1}^{N} u\big(\cls\, e_m(x^{m}_{i,v})\big) \big\rVert},
  \qquad u(c) = \frac{\LN(c)}{\lVert \LN(c) \rVert},
\end{equation}
where $\LN$ denotes a layer normalisation without parameters. The translation loss in the space of $m$ is an InfoNCE objective
over the global batch,
\begin{equation}
  \Lc^{m}_{\mathrm{CLS}} = -\frac{1}{BN}\sum_{i=1}^{B}\sum_{v=1}^{N}
  \log \frac{\exp\big(\langle q^{m}_{i,v}, c^{m}_i\rangle / T\big)}{\sum_{j=1}^{B} \exp\big(\langle q^{m}_{i,v}, c^{m}_j\rangle / T\big)},
  \label{eq:clsnce}
\end{equation}
with temperature $T = 0.1$; the two spaces contribute $\Lc^{a}_{\mathrm{CLS}} + \Lc^{b}_{\mathrm{CLS}}$. The objective requires the
translated embedding of a sample to be closer to the own embedding of the same sample in the target space than to the own embedding
of any other sample in the batch. Every vector in Eq.~\eqref{eq:clsnce} lies in the space of $m$: the own embeddings of $\om$ do
not appear, so no loss term compares the two own spaces directly. Neither side is excluded from the gradient. The target centroid
$c^{m}_i$ is therefore also optimised by the translation loss, which couples the own space of $m$ to the translation from $\om$;
the two encoders are trained jointly, not independently. A detached target, as used in predictive formulations
\cite{assran2023ijepa,levljepa}, is evaluated in Section~\ref{sec:ablation}. The candidates $c^{m}_j$ are gathered from all devices
with their gradient, so that the negatives of an anchor comprise all other samples of the global batch.

\subsection{Skipless translator}
\label{sec:skipless}

The layers of the image towers are pre-normalisation transformer layers with learned scale vectors $\gamma_1, \gamma_2$, and those
of the text tower are post-normalisation BERT layers:
\begin{align}
  \text{image (pre-norm):}\quad & y = x + \gamma_1 \odot \operatorname{Attn}\big(\LN_1(x)\big), &
  x' &= y + \gamma_2 \odot \operatorname{MLP}\big(\LN_2(y)\big), \label{eq:residual} \\
  \text{text (post-norm):}\quad & y = \LN_1\big(x + \operatorname{Attn}(x)\big), &
  x' &= \LN_2\big(y + \operatorname{MLP}(y)\big). \nonumber
\end{align}
A translator built from such a layer initially outputs the own embedding of the tower plus a small update. Eq.~\eqref{eq:clsnce}
can then be reduced either by learning a mapping from one space to the other or by moving the own embeddings of the two towers
towards each other until the identity itself acts as a translation; in our experiments the residual translator led to the second
outcome (Section~\ref{sec:ablation}). We therefore remove both residual additions in the translator layer, and only there:
\begin{align}
  \text{image:}\quad & y = \gamma_1 \odot \operatorname{Attn}\big(\LN_1(x)\big), & x' &= \gamma_2 \odot \operatorname{MLP}\big(\LN_2(y)\big),
  \label{eq:skipless} \\
  \text{text:}\quad & y = \LN_1\big(\operatorname{Attn}(x)\big), & x' &= \LN_2\big(\operatorname{MLP}(y)\big). \nonumber
\end{align}
The skipless layer provides no path through which the source embedding can be passed on unchanged, so that any correspondence
between the spaces must be computed by the translator. A transformer layer without residual connections loses its signal under a
standard initialisation \cite{noci2022signal,he2023deep}; we re-initialise the translator layer following \citet{ji2025skipless},
with orthogonal value, output and MLP weights and a query-key product close to a scaled identity (Appendix~\ref{app:impl}). The
skipless layer and its initialisation are those of \citet{ji2025skipless}; the use of such a layer as a cross-modal translator, and
its effect on the alignment of the encoders, is what we study here. We use a single translator layer; deeper skipless translators
did not improve the results in our ablations (Section~\ref{sec:ablation}).

\subsection{Patch-level translation}
\label{sec:patch}

When both modalities are images acquired on a common pixel grid, as the optical and radar images of
Section~\ref{sec:data-eo}, the translation can be required of every patch and not only of the class token. The direct
formulation, which matches each translated patch to the target patch at the same position, admits a shortcut: when patches
correspond only approximately, a pattern that depends on position alone and is shared by all images reduces a position-matched
cost without encoding any content. We instead require each translated patch to identify the correct \emph{image} at its position
(Figure~\ref{fig:patch}). Let the rows $r$ index the global views of the batch, and let $t_{r,k}$ be the translated and $o_{r,k}$ the
own patch token at position $k$ of row $r$, both normalised by $u$ and with the register tokens excluded. With $\operatorname{sg}$
the stop-gradient operator,
\begin{equation}
  \Lc^{m}_{\mathrm{patch}} = -\frac{1}{|\mathcal{V}|\,P}\sum_{r \in \mathcal{V}}\sum_{k=1}^{P}
  \log \frac{\exp\big(\langle t_{r,k}, \operatorname{sg}[o_{r,k}]\rangle / T\big)}
  {\sum_{r' \in \mathcal{R}(r)} \exp\big(\langle t_{r,k}, \operatorname{sg}[o_{r',k}]\rangle / T\big)},
  \label{eq:patchnce}
\end{equation}
where $\mathcal{V}$ is the set of global views, $P = 256$ the number of patches, and $\mathcal{R}(r)$ contains row $r$ and every row
of the other samples of the global batch, gathered from all devices without gradient; the other global view of the same sample is
excluded. A component shared by all images at a given position increases the positive and every negative similarity equally and
therefore does not reduce the loss; only content that discriminates between images does. The own patches are detached, so that
the term shapes the translating tower and leaves the target space to its own objective. The term is defined on representations:
it asks the translated patch grid to match the target's own patch embeddings, not to reconstruct the target image.

\begin{figure}[!t]
  \centering
  \resizebox{\linewidth}{!}{\input{duojepa_patch_nce_figure_generic.tex}}
  \caption{The patch InfoNCE in the space of a target modality $m$ pulls a translated patch of the source modality $\bar m$
  towards the detached own patch of the same image at the same position and pushes it away from that position in the other
  images of the batch.}
  \label{fig:patch}
\end{figure}

\subsection{Training objective and schedule}
\label{sec:total}

The complete objective combines the self-supervised loss of each tower with the translation terms of both spaces,
\begin{equation}
  \Lc = \sum_{m \in \{a, b\}} \Lc^{m}_{\mathrm{LeJEPA}}
  + \rho(t) \sum_{m \in \{a, b\}} \Big[ \Lc^{m}_{\mathrm{CLS}} + \Lc^{m}_{\mathrm{patch}} \Big],
  \qquad
  \rho(t) = \clip\!\Big(\frac{t - W}{rW - W},\, 0,\, 1\Big),
  \label{eq:total}
\end{equation}
where $\Lc^{m}_{\mathrm{patch}}$ is included only for images acquired on a common pixel grid, $t$ is the training step and $W$ the
length of the learning-rate warm-up. The translation is activated after the warm-up and reaches its full weight at $rW$ ($r = 2$
for the satellite images and $1.5$ for radiographs and reports), so that each encoder forms its own space before the translation
loss acts on it. The translator layers receive no gradient before $W$; their learning rate then follows its own warm-up of $W$
steps and a cosine decay. Appendix~\ref{app:impl} lists the remaining settings.

\section{Experimental Setup}
\label{sec:eval}

\paragraph{Models.} For radiographs and reports, the image tower is a ViT-B/14 with four register tokens
\cite{dosovitskiy2021vit,oquab2024dinov2,darcet2024registers} trained from random initialisation, and the text tower a BERT-base
model pre-trained on biomedical text \cite{gu2021pubmedbert}; the model is trained once (seed 42) for 20k steps with a batch of 128
pairs. For satellite images, both towers are ViT-B/14 networks trained from random initialisation, one operating on the 12 optical
bands and one on the two radar polarisations, once (seed 42) for 100k steps with a batch of 256 pairs. Table~\ref{tab:recipes} in
the appendix lists all settings.

\paragraph{Baselines.} The comparisons are of two kinds, which should be read differently. \emph{Reproduced under our protocol}:
LeVLJEPA \cite{levljepa} is reimplemented and trained by us on the same pre-training images and reports, with the same two encoder
bodies as DuoJEPA (the ViT-B/14 image encoder from random initialisation and the biomedical BERT text encoder), its pre-projection
and predictor heads, its mean-squared prediction loss with stop-gradient targets and SIGReg on both sides (weight 0.01 each), one
view per modality, a batch of 256 pairs and 20k steps (seed 42). It differs from the published recipe in the encoders (the authors
use a ViT-B/16 and a GPT-2 text model trained from scratch), the input normalisation and the learning rate ($2\times10^{-4}$ instead
of $4\times10^{-2}$, which did not train stably in our setting); its evaluation reads the pre-projected class token as the own
embedding and the predictor output as the translated embedding. CheXzero \cite{tiu2022chexzero} and GLoRIA \cite{huang2021gloria}
are evaluated with our code from their released checkpoints. CheXzero was trained on 377,110 image-report pairs of MIMIC-CXR; our
code reproduces its published zero-shot AUC with its released ensemble (88.94 against 88.9), and we report its single best
checkpoint. GLoRIA was trained on 191,229 frontal image-report pairs of the CheXpert training set; our linear probe reproduces its
published values within 0.22 points. \emph{Reported from the literature}: the Sentinel foundation models of Table~\ref{tab:vv-main}
are not re-evaluated by us; their values are taken from Copernicus-Bench \cite{wang2025copernicusfm}, except those of TerraFM, which
come from its own paper \cite{danish2025terrafm}. They differ from DuoJEPA in pre-training data, architecture, parameter count and
readout, and our implementation of the protocol differs from the official code (Appendix~\ref{app:eval}), so that this comparison
is not controlled.

\paragraph{Evaluation metrics.} The evaluation addresses four properties of each model, each measured with an established or
explicitly defined protocol. All AUC values and percentages are reported on a 0--100 scale.
\begin{itemize}
  \item \emph{Quality of the own embeddings.} A linear probe on the frozen own embedding of an image tower predicts the labels of
        each dataset. For radiographs we report the mean area under the receiver operating characteristic curve (AUC) over the
        five findings, trained with 1, 10 and 100\% of the training labels (five label draws for 1 and 10\%, one fit for 100\%)
        and tested on the 202 validation images. The AUC equals the probability that a randomly chosen positive image is scored
        above a randomly chosen negative one (50 corresponds to chance). For BigEarthNet we follow the Copernicus-Bench protocol
        \cite{wang2025copernicusfm}: 11,894 training, 6,117 validation and 5,991 test images, a linear classifier trained for 50
        epochs, three runs with different subsamples (seeds 0, 1, 2), and the micro-averaged mean average precision (mAP), which
        ranks all image-class decisions by the classifier score and measures how early the correct decisions appear.
  \item \emph{Cross-modal transfer.} For radiographs, \emph{zero-shot classification} compares each test image with the prompts
        ``$\langle$finding$\rangle$'' and ``no $\langle$finding$\rangle$'' following CheXzero \cite{tiu2022chexzero}, as the mean
        AUC over the five findings, and the 5x200 accuracy and the precision of the five retrieved reports (P@5) follow GLoRIA
        \cite{huang2021gloria}. Retrieval is measured by the recall at 10 (R@10), the percentage of queries whose partner is among
        the ten most similar candidates, on the 202 validation pairs (chance 4.95). The \emph{cross-modal score} is the mean of
        the image-to-report and report-to-image recall in the image space and the image-to-report recall in the text space. For
        satellite images we report the top-10 retrieval of the paired image among 4,000 test images (chance 0.25), and
        \emph{substitution}, defined below.
  \item \emph{Direct alignment of the own spaces.} For radiographs we compute, on the 1,859 held-out pairs and without any fitted
        mapping, the AUC with which the cosine of a pair exceeds the cosines of the same image with the 1,858 other reports, after
        centring each modality, averaged over images. A permutation test that shuffles the pairing 200 times gives a p-value
        (smallest attainable value 0.005). Following the terminology of \citet{maniparambil2024vision}, we call two spaces aligned
        when the AUC exceeds 55 with $p < 0.01$, and unaligned otherwise. An AUC near 50 indicates that this test detects no direct
        cosine alignment; it does not show that the two spaces are unrelated, and Section~\ref{sec:rotation} shows that they are
        related by an approximate rotation. The 1,859 pairs come from 645 patients and are therefore not independent; its 95\%
        confidence interval is obtained by a bootstrap over patients. For satellite images we report the top-10 retrieval between the two own spaces without
        the translator.
  \item \emph{Spatial fidelity of the translation.} For satellite images the translated patch grid is compared with the own grid
        of the target. Let $q = \frac{1}{P}\sum_k [\cos(t_k, o_k) - \cos(t_k, o'_k)]$, averaged over images, where $o'_k$ is the
        patch at position $k$ of another image; subtracting the same-position cosine of another image removes any
        position-only component. The \emph{patch translation score} is
        $(q_{\mathrm{trans}} - q_{\mathrm{floor}}) / (q_{\mathrm{ceil}} - q_{\mathrm{floor}}) \times 100$, where the floor repeats the
        translated class token at every position and the ceiling uses the target's own grid. The \emph{within-image structure} is
        the Pearson correlation between the upper triangles of the $256 \times 256$ patch-similarity matrices of the translated
        grid and of the source grid of the same image, averaged over images; the same quantity computed against another image's
        source grid serves as a baseline. Both measures concern representations; neither measures the reconstruction of the
        target image.
\end{itemize}
\emph{Substitution} keeps a linear classifier fitted on the own embeddings of one sensor and applies it, on the same test images,
to three inputs: the genuine own embeddings of that sensor (ceiling, $s_{\mathrm{ceil}}$), the translations of the other sensor's
images into that space ($s_{\mathrm{trans}}$), and the untranslated own embeddings of the other sensor (floor, $s_{\mathrm{floor}}$).
We report $(s_{\mathrm{trans}} - s_{\mathrm{floor}}) / (s_{\mathrm{ceil}} - s_{\mathrm{floor}}) \times 100$ together with the three
micro mAP values; the share is not reported when $s_{\mathrm{ceil}} - s_{\mathrm{floor}} < 5$ points. The share is a fraction of a
classification-performance gap; it does not measure how much of the missing sensor's information is reconstructed. All models are
evaluated through their standard forward pass, without test-time modification. For the radiograph-report models we report 95\%
bootstrap confidence intervals over the test samples (2,000 resamples; Appendix~\ref{app:ci}); they reflect the variability of the
evaluation sets, not that of training. Appendix~\ref{app:eval} details the protocols.

\section{Results}
\label{sec:results}

\subsection{Chest radiographs and reports}
\label{sec:vl-results}

\begin{table}[!t]
\centering\small
\caption{Chest radiograph--report results, with DuoJEPA and the reimplemented LeVLJEPA trained on CheXpert Plus and two aligned
contrastive models evaluated from their released checkpoints.}
\label{tab:vl-main}
\begin{adjustbox}{max width=\textwidth}
\begin{tabular}{lccccccccc}
\toprule
 & & \multicolumn{5}{c}{cross-modal (image space unless noted)} & & \multicolumn{2}{c}{own embeddings} \\
\cmidrule(lr){3-7}\cmidrule(lr){9-10}
model & params (M) & zero-shot AUC img\,/\,txt & 5x200 acc\,/\,P@5 & R@10 i$\to$t\,/\,t$\to$i & R@10 text sp. & cross-modal & & probe AUC 1\,/\,10\,/\,100\% & alignment AUC ($p$) \\
\midrule
\B{DuoJEPA} & 208.8 & \B{\rr{86.6}{86.3}} & \rr{47.5}{56.0} & \rr{78.2}{78.7} & \B{78.7} & \B{78.5} & & 81.0\,/\,85.2\,/\,86.7 & 48.3 (0.995), unaligned \\
LeVLJEPA (reimpl.) \cite{levljepa} & 232.4 & \rr{60.0}{62.6} & \rr{39.9}{49.8} & \rr{63.4}{62.4} & 66.8 & 64.2 & & 84.7\,/\,86.4\,/\,86.8 & 51.0 (0.065), unaligned \\
\midrule
CheXzero \cite{tiu2022chexzero} & 151.3 & 88.2 & \rr{46.3}{48.3} & \rr{41.1}{42.1} & -- & -- & & 85.0\,/\,86.2\,/\,86.6 & 88.7 (0.005), aligned \\
GLoRIA \cite{huang2021gloria} & 134.2 & 64.7 & \rr{58.8}{60.9} & \rr{86.1}{84.7} & -- & -- & & 86.8\,/\,87.6\,/\,88.0 & 98.9 (0.005), aligned \\
\bottomrule
\end{tabular}
\end{adjustbox}
\end{table}

The columns of Table~\ref{tab:vl-main} are defined as follows. \emph{Params} is the number of parameters, in millions, of the
image and text encoders used at inference, including the two translator layers of DuoJEPA and the two pre-projection and
predictor networks of LeVLJEPA and excluding projectors used only during training; all counts were computed from the evaluated
checkpoints. \emph{Zero-shot AUC} is the mean AUC over the five findings of the CheXpert test set, computed in the image space
and, after the slash, in the text space. \emph{5x200 acc\,/\,P@5} denote the five-class zero-shot accuracy on CheXpert 5x200
(chance 20) and the precision of the five reports retrieved for each image. \emph{R@10 i$\to$t\,/\,t$\to$i} is the recall at 10
for image-to-report and report-to-image retrieval among the 202 validation pairs in the image space, \emph{R@10 text sp.} the
image-to-report recall in the text space, and \emph{cross-modal} the mean of these three values. \emph{Probe AUC} is the mean AUC
of a linear classifier on the frozen image embedding. \emph{Alignment AUC} is the paired-cosine AUC between the two own spaces with
its permutation p-value. For the aligned models the two spaces coincide, so that a single value is reported and the text-space
columns do not apply. The DuoJEPA and LeVLJEPA rows are single training runs.

DuoJEPA outperforms our reimplementation of LeVLJEPA on every cross-modal measure: the zero-shot AUC increases from 60.0 to 86.6,
the five-class accuracy from 39.9 to 47.5, and the cross-modal retrieval score from 64.2 to 78.5, with fewer parameters at inference
(208.8M against 232.4M), since the two translator layers of DuoJEPA (14.2M) are smaller than the pre-projection and predictor
networks of LeVLJEPA (37.8M) that operate on the same encoder bodies. The paired bootstrap intervals of the differences exclude zero
for every cross-modal measure, for instance $+26.6$ [22.8, 30.4] for the zero-shot AUC and $+14.4$ [8.6, 20.0] for the cross-modal
score (Appendix~\ref{app:ci}). The improvement is not obtained by directly aligning the two modalities: the paired-cosine test detects no
alignment between the own spaces of DuoJEPA (AUC 48.3, 95\% interval [46.1, 50.3], $p = 0.995$), as for LeVLJEPA (51.0,
$p = 0.065$) and in contrast to CheXzero (88.7, $p = 0.005$) and GLoRIA (98.9, $p = 0.005$; its value may be raised by the fact that
the hold-out pairs come from CheXpert training patients, on which GLoRIA was trained). The image and the report of a pair are therefore not closer in either own space than other pairs, and
the cross-modal performance is obtained through the learned translation.

The aligned contrastive models delineate the trade-offs of the approach. CheXzero, trained on a different and larger corpus,
attains a zero-shot AUC 1.6 points higher, within overlapping intervals (88.2 [86.5, 89.9] against 86.6 [84.8, 88.3]), but a
substantially lower retrieval. GLoRIA, which adds a local alignment term between
image regions and words, achieves higher retrieval, also on the validation pairs, which are its own held-out test set. It also leads
on 5x200; since GLoRIA was pre-trained on the CheXpert training set from which the 5x200 images are drawn, part of this advantage
may reflect overlap with its training data, which we could not verify because its draw is not released. Its zero-shot AUC under
the CheXzero prompts is 64.7, but its own prompt processing differs from that protocol. Under the linear probe, the image tower of
LeVLJEPA is the stronger representation in the label-efficient regime: 84.7 $\pm$ 0.7 against 81.0 $\pm$ 0.9 with 1\% of the labels
and 86.4 $\pm$ 0.3 against 85.2 $\pm$ 0.6 with 10\% (mean $\pm$ standard deviation over five label draws), and 86.8 against 86.7
with all labels. DuoJEPA thus obtains its cross-modal capability without directly aligned own spaces, at a moderate cost in the
label-efficient regime of the image encoder.

\subsection{Optical and radar satellite images}
\label{sec:vv-results}

\begin{table}[!t]
\centering\small
\caption{Linear-probe micro mAP on BigEarthNet v2, with DuoJEPA evaluated under our reproduction of the Copernicus-Bench protocol
and published Sentinel foundation models as reported in the literature.}
\label{tab:vv-main}
\begin{adjustbox}{max width=\textwidth}
\begin{tabular}{lllcc}
\toprule
model & backbone, params (M) & pre-training data & BE-S1 & BE-S2 \\
\midrule
\multicolumn{5}{l}{\emph{evaluated by us}} \\
\B{DuoJEPA}, own CLS & ViT-B/14, 86.4\,/\,87.9 & SSL4EO-S12 & 76.7 $\pm$ 0.1 & 81.7 $\pm$ 0.3 \\
\B{DuoJEPA}, own mean-pooled patches & ViT-B/14, 86.4\,/\,87.9 & SSL4EO-S12 & 77.1 & 81.8 \\
\B{DuoJEPA}, own + translated CLS & ViT-B/14 + translator, 93.5\,/\,95.0 & SSL4EO-S12 & 77.4 & 81.1 \\
\midrule
\multicolumn{5}{l}{\emph{reported in the literature (not controlled)}} \\
TerraFM \cite{danish2025terrafm} & ViT-B/16, 120 & own corpus & 76.9 & 84.4 \\
SoftCon \cite{wang2024softcon} & ViT-B/14, 86 & own corpus & 78.7 & 83.6 \\
Copernicus-FM \cite{wang2025copernicusfm} & ViT-B/16, 113.1 & Copernicus-Pretrain & 77.9 & 79.0 \\
CROMA \cite{fuller2023croma} & ViT-B/8, 42.6\,/\,85.6 & SSL4EO-S12 & 70.8 & 76.4 \\
DOFA \cite{xiong2024dofa} & ViT-B/16, 111.3 & own corpus & 70.5 & 75.5 \\
supervised & ViT-B/16, 85.6\,/\,87.6 & -- & 70.6 & 80.1 \\
supervised & ViT-S/16, 21.6\,/\,22.6 & -- & 71.5 & 79.5 \\
random initialisation & ViT-B/16, 85.6\,/\,87.6 & -- & 63.8 & 71.6 \\
\bottomrule
\end{tabular}
\end{adjustbox}
\end{table}

\begin{table}[!t]
\centering\small
\caption{Substitution of a missing sensor on BigEarthNet v2, with the micro mAP of floor, translated input and ceiling and the
share of the gap closed, and the additional probe settings of CROMA in micro mAP.}
\label{tab:vv-subst}
\begin{adjustbox}{max width=\textwidth}
\begin{tabular}{lcccc}
\toprule
 & \multicolumn{2}{c}{class token} & \multicolumn{2}{c}{mean-pooled patches} \\
\cmidrule(lr){2-3}\cmidrule(lr){4-5}
substitution & floor / translated / ceiling & share & floor / translated / ceiling & share \\
\midrule
optical classifier, radar image translated & 15.96 / 65.24 / 81.72 & 74.9\% & 16.32 / 55.28 / 81.81 & 59.5\% \\
radar classifier, optical image translated & 18.65 / 75.55 / 76.68 & 98.0\% & 16.45 / 64.20 / 77.06 & 78.8\% \\
\midrule
additional probe settings & \multicolumn{2}{c}{class token} & \multicolumn{2}{c}{mean-pooled patches} \\
\midrule
radar + optical, own embeddings concatenated & \multicolumn{2}{c}{81.0} & \multicolumn{2}{c}{80.6} \\
1\% of the labels: S1\,/\,S2\,/\,S1 + S2 & \multicolumn{2}{c}{74.4\,/\,78.3\,/\,78.0} & \multicolumn{2}{c}{74.6\,/\,78.7\,/\,77.6} \\
MLP probe: S1\,/\,S2\,/\,S1 + S2 & \multicolumn{2}{c}{77.9\,/\,81.8\,/\,81.7} & \multicolumn{2}{c}{78.3\,/\,82.5\,/\,82.8} \\
\bottomrule
\end{tabular}
\end{adjustbox}
\end{table}

In Table~\ref{tab:vv-main}, BE-S1 and BE-S2 denote the linear-probe micro mAP on the Sentinel-1 (radar) and Sentinel-2 (optical)
images of BigEarthNet, and the second column gives the backbone and the parameters, in millions, of the encoder probed (radar\,/\,
optical where the two differ). For DuoJEPA the radar column is obtained from the radar tower and the optical column from the
optical tower, with the standard deviation over the three probe runs; \emph{own CLS} is the class token of the own embedding,
\emph{own mean-pooled patches} the average of its patch tokens, and \emph{own + translated CLS} the concatenation of the own class
token with the translation of the same image into the space of the other sensor, so that only one sensor is required. The values
of the published models are those of Table~4 of Copernicus-Bench \cite{wang2025copernicusfm}, except for TerraFM, whose values come
from its own paper, in which the probe additionally searches over the feature layer and the readout. The parameter counts of
SoftCon and TerraFM are taken from their papers; all other counts were computed by us, those of Copernicus-FM, CROMA and DOFA from
their released checkpoints. For CROMA they cover the unimodal encoder of each sensor, to which its joint radar-optical encoder adds
56.7M; for DOFA and Copernicus-FM they include the wavelength-conditioned patch embedding shared by all sensors, while the branch
of Copernicus-FM for non-spectral variables (26.4M) is excluded.

The own class token of DuoJEPA reaches 76.7 micro mAP on radar images, comparable to TerraFM (76.9) and below
SoftCon (78.7) and Copernicus-FM (77.9), and 81.7 on optical images, above Copernicus-FM and the supervised baselines and below
TerraFM (84.4) and SoftCon (83.6). CROMA, the only listed model pre-trained on the same SSL4EO-S12 corpus, is reported at 70.8 and
76.4. Mean pooling of the own patch tokens yields slightly higher values (77.1 and 81.8). Concatenating the own embedding with the
translation of the same image improves the radar result (77.4) but not the optical one (81.1), and the concatenation of the own
embeddings of both sensors (81.0, Table~\ref{tab:vv-subst}) does not exceed the optical embedding alone under the linear probe;
a nonlinear probe on both reaches 81.7 with the class tokens and 82.8 with the mean-pooled patches. Our implementation of the probe carries, beyond the variation over its three runs,
an additional variability of about $\pm$0.5 mAP from the random initialisation of the classifier (Appendix~\ref{app:eval}), so that
differences of this order to published values are not interpretable.

Table~\ref{tab:vv-subst} reports, in its upper part, the substitution scores with the micro mAP of floor, translated input and
ceiling, and in its lower part the additional probe settings of CROMA on the same subsamples: the own embeddings of both sensors
concatenated, linear probes trained on 1\% of the clean training split (2,379 images), and a nonlinear probe with one hidden layer
of 2,048 units. The columns indicate the readout, either the class token or the mean of the patch tokens.

The translation is accurate at the level of the class token and of the patch grid. Through the translator, the paired image is
among the ten nearest of 4,000 candidates in 77.3\% of cases from optical to radar and in 79.2\% from radar to optical (chance
0.25\%), whereas the two own spaces retrieve each other's images at 0.45\%. At the patch level, the translated grid closes 69 and
68\% of the gap between a grid that repeats the class token and the own grid of the target, and its patch-similarity structure
correlates with that of the source image at 0.95 and 0.94, against approximately 0.5 for an unrelated image. The translation
therefore transfers within-image similarity structure in addition to global content; this is a statement about representations
and not about the reconstruction of the target image. In terms of substitution, a radar classifier applied to translated optical
images reaches 75.55 mAP, against 76.68 on genuine radar images and 18.65 on the untranslated optical embeddings, that is 98.0\% of
the gap. In the opposite direction, an optical classifier applied to translated radar images reaches 65.24 against 81.72 and 15.96,
or 74.9\%. A plausible explanation of this asymmetry, not tested here, is that optical images carry more of the information
relevant to land-cover recognition than radar images (81.7 against 76.7 mAP), and a translation cannot supply information absent
from its source. Mean-pooled translated patches substitute less effectively than the translated class token (78.8 and
59.5\%).

\subsection{Ablation study}
\label{sec:ablation}

\begin{table}[!t]
\centering\small
\caption{Single-change ablations of the chest radiograph--report recipe and the steps from the ablation base to the final recipe
(one training run per row, except the three seeds of the base).}
\label{tab:vl-abl}
\begin{adjustbox}{max width=\textwidth}
\begin{tabular}{lcccccccc}
\toprule
 & \multicolumn{5}{c}{cross-modal (image space unless noted)} & & \multicolumn{2}{c}{own embeddings} \\
\cmidrule(lr){2-6}\cmidrule(lr){8-9}
run & zero-shot AUC img\,/\,txt & 5x200 acc\,/\,P@5 & R@10 i$\to$t\,/\,t$\to$i & R@10 text sp. & cross-modal & & probe AUC 1\,/\,10\,/\,100\% & alignment AUC ($p$) \\
\midrule
base, seed 42 & \rr{83.0}{83.2} & \rr{46.3}{54.6} & \rr{80.7}{76.7} & 81.2 & 79.5 & & 79.2\,/\,83.9\,/\,87.3 & 50.8 (0.070) \\
base, seed 43 & \rr{84.3}{84.0} & \rr{45.3}{55.6} & \rr{79.2}{78.7} & 76.7 & 78.2 & & 79.4\,/\,83.7\,/\,86.2 & 50.3 (0.373) \\
base, seed 44 & \rr{83.0}{82.9} & \rr{46.7}{55.6} & \rr{76.7}{77.2} & 75.7 & 76.6 & & 80.6\,/\,84.2\,/\,87.9 & 54.7 (0.005) \\
\midrule
own side detached & \rr{81.1}{77.0} & \rr{41.0}{43.1} & \rr{55.4}{55.4} & 19.3 & 43.4 & & 81.6\,/\,85.3\,/\,86.0 & 53.2 (0.005) \\
$T$ 0.05 & \rr{85.1}{84.1} & \rr{41.8}{54.9} & \rr{79.7}{80.2} & 79.7 & 79.9 & & 78.2\,/\,84.5\,/\,86.3 & 51.5 (0.015) \\
$T$ 0.2 & \rr{79.6}{80.3} & \rr{45.4}{50.7} & \rr{74.3}{74.8} & 74.8 & 74.6 & & 75.7\,/\,81.9\,/\,84.7 & 52.3 (0.005) \\
2 skipless layers & \rr{82.8}{82.5} & \rr{45.5}{55.2} & \rr{77.7}{78.7} & 77.2 & 77.9 & & 78.9\,/\,83.4\,/\,85.2 & 47.7 (1.000) \\
3 skipless layers & \rr{83.8}{83.7} & \rr{45.2}{54.7} & \rr{78.7}{79.2} & 78.2 & 78.7 & & 79.3\,/\,83.4\,/\,87.5 & 51.0 (0.075) \\
1 \emph{residual} layer & \rr{83.6}{83.4} & \rr{43.4}{55.0} & \rr{77.2}{77.2} & 77.2 & 77.2 & & 80.0\,/\,84.0\,/\,85.4 & \B{97.5 (0.005), aligned} \\
batch 128 & \rr{85.2}{85.4} & \rr{48.5}{55.1} & \rr{77.2}{78.7} & 77.7 & 77.9 & & 80.8\,/\,85.0\,/\,86.2 & 51.7 (0.010) \\
\midrule
\multicolumn{9}{l}{\emph{from batch 128 to the final recipe: every batch-coupled term computed over the global batch}} \\
InfoNCE candidates, SIGReg, BatchNorm global & \rr{86.0}{86.2} & \rr{47.1}{55.5} & \rr{81.2}{77.7} & 80.2 & 79.7 & & 81.1\,/\,85.7\,/\,86.6 & 50.4 (0.289) \\
\quad + text invariance global (\B{final}) & \B{\rr{86.6}{86.3}} & \rr{47.5}{56.0} & \rr{78.2}{78.7} & 78.7 & 78.5 & & 81.0\,/\,85.2\,/\,86.7 & 48.3 (0.995) \\
\quad SIGReg $\lambda$ halved instead & \rr{86.2}{86.6} & \rr{47.7}{56.4} & \rr{81.2}{80.2} & 80.7 & 80.7 & & 80.5\,/\,85.4\,/\,87.2 & \B{55.4 (0.005), aligned} \\
\bottomrule
\end{tabular}
\end{adjustbox}
\end{table}

\begin{table}[!t]
\centering\small
\caption{Ablations of the satellite recipe at 20k steps, each changing one setting except the batch-size run and the control
without translation (one training run per row, except the two seeds of the base), with pairs of values given as optical to radar\,/\,radar to optical except the probe.}
\label{tab:vv-abl}
\begin{adjustbox}{max width=\textwidth}
\begin{tabular}{lcccccc}
\toprule
run & probe S1\,/\,S2 (mAP) & retrieval top-10 & patch translation & within-image structure & own$\leftrightarrow$own & substitution \\
\midrule
base & \rr{73.79}{80.84} & \rr{65.8}{66.8} & \rr{19}{29}\% & \rr{0.89}{0.89} & 0.23\% & \rr{92.1}{65.6}\% \\
seed 43 & \rr{73.64}{80.75} & \rr{67.2}{67.9} & \rr{21}{31}\% & \rr{0.90}{0.89} & 0.53\% & \rr{92.0}{69.3}\% \\
no translation, $\lambda$ 0.05 & \rr{73.71}{80.53} & -- & -- & -- & 0.13\% & -- \\
no patch term & \rr{73.39}{80.80} & \rr{63.1}{65.5} & \rr{$-1$}{1}\% & \rr{0.65}{0.79} & 0.48\% & \rr{91.5}{67.5}\% \\
2 skipless layers & \rr{73.89}{80.72} & \rr{63.9}{65.6} & \rr{15}{22}\% & \rr{0.83}{0.83} & 0.35\% & \rr{94.7}{64.0}\% \\
1 \emph{residual} layer & \rr{72.60}{79.72} & \rr{64.5}{65.0} & \rr{$-4$}{0}\% & \rr{1.00}{1.00} & \B{64.7\%} & \rr{n/a}{$-11.5$}\% \\
$T$ 0.05 & \rr{73.75}{81.07} & \rr{67.0}{69.1} & \rr{9}{18}\% & \rr{0.89}{0.87} & 0.35\% & \rr{80.1}{37.3}\% \\
$T$ 0.2 & \rr{73.02}{79.81} & \rr{49.2}{52.3} & \rr{44}{50}\% & \rr{0.92}{0.89} & 0.18\% & \rr{97.5}{67.7}\% \\
batch 128, $\lambda$ 0.01 & \rr{72.31}{79.93} & \rr{54.0}{56.5} & \rr{32}{30}\% & \rr{0.89}{0.87} & 0.25\% & \rr{90.0}{59.3}\% \\
SIGReg $\lambda$ 0.01 & \rr{73.73}{80.63} & \rr{66.3}{67.0} & \rr{21}{30}\% & \rr{0.91}{0.90} & 0.13\% & \rr{94.9}{64.9}\% \\
\bottomrule
\end{tabular}
\end{adjustbox}
\end{table}

Each component of the recipe is varied in isolation, with 20k-step runs in both applications. Apart from the seed repetitions of
the two base configurations, every row of Tables~\ref{tab:vl-abl} and~\ref{tab:vv-abl} is a single training run, and no test of
statistical significance is applied; the differences are interpreted against the variation between base seeds. The radiograph-report
ablations (Table~\ref{tab:vl-abl}) use the columns of Table~\ref{tab:vl-main}. Their base is the recipe of
Section~\ref{sec:method} with a batch of 256 and per-device batch statistics; \emph{own side detached} applies a stop-gradient to
the target centroid $c^{m}_i$ of Eq.~\eqref{eq:clsnce}, the following rows change the temperature, the number of skipless
translator layers, the translator type and the batch size, and the lower block lists the steps from the batch-128 configuration to
the final model. The range of the three base seeds on the cross-modal score, 2.9 points, serves as the margin within which
differences are not interpreted.

The satellite ablations (Table~\ref{tab:vv-abl}) report, from left to right, the linear-probe micro mAP of the own class token of
the radar and the optical tower (mean over three probe runs; standard deviations between 0.09 and 0.55); the top-10 retrieval of the paired image through the translator among 4,000 test images; the patch
translation score; the within-image structure; the top-10 retrieval between the two own spaces without the translator
(\emph{own$\leftrightarrow$own}, chance 0.25\%, high values indicating direct alignment); and the class-token substitution share.
All pairs except the probe are ordered optical to radar\,/\,radar to optical; for substitution, optical to radar denotes the radar
classifier applied to translated optical images. An entry n/a indicates that the gap between floor and ceiling is below 5 points.
The two seeds of the base differ by at most 0.15 mAP on the probe, 1.4 points on retrieval, 2 points on patch translation and 3.7
points on substitution; we interpret differences beyond 0.5 mAP on the probe, 3 points on retrieval and patch translation, and 4
points on substitution.

\paragraph{Removing the translation objective.} Without the translation terms, the two towers are independent LeJEPA encoders.
This control was trained with a SIGReg weight of 0.05 instead of 0.005, so that it differs from the base in two settings. Its
translator layers receive no gradient, and the translation columns of Table~\ref{tab:vv-abl} are therefore not defined for it;
measured through the untrained layers, the paired image is retrieved at chance level (0.1 and 0.2\% top-10). The own class tokens
are probed at 73.7 and 80.5 mAP, against 73.8 and 80.8 for the base, and at 74.9 and 80.9 against 74.6 and 81.4 with per-feature
standardisation, all within the margin. The two own spaces retrieve each other's images in 0.13\% of cases (chance 0.25\%), as
unaligned as in the base. In this setting the translation objective thus leaves the class-level quality of the own embeddings
unchanged, as measured by the linear probe, and adds the translation; Section~\ref{sec:rotation} shows that it also establishes a
near-rotational relation between the two own spaces that the independently trained encoders share to a much smaller extent.

\paragraph{Residual versus skipless translator.} Replacing the skipless layer by a residual one leaves the cross-modal scores close
to the base, but the two encoders become directly aligned in both applications. For radiographs and reports, the alignment AUC rises
to 97.5 ($p = 0.005$), against 47.7 to 55.4 for the skipless configurations, of which only one (the halved SIGReg weight, 55.4)
meets the aligned criterion, and the own spaces retrieve each other with an R@10 of
66.3 on held-out halves of 930 pairs (chance 1.08\%), the value reached by a rotation fitted between them (66.0). For
satellite images, the own spaces retrieve each other's images in 64.7\% of cases, against at most 0.53\% for every skipless run,
and the translated grid reproduces the structure of the source grid exactly (correlation 1.00) while the patch translation score
falls to zero. The residual run also loses approximately one point on both probes (72.6 and 79.7 against 73.8 and 80.8), and
substitution is not defined, since the untranslated embedding already performs nearly as well as the translated one. In this
setting, the residual connection did not improve the translation but led the encoders to absorb it. Deeper skipless
translators remained within the margin for radiographs (two and three layers), and two layers reduced the patch translation for
satellite images (three were not tested there).

\paragraph{Gradient through the target embedding.} Detaching the own side of Eq.~\eqref{eq:clsnce}, as in predictive
formulations, preserved class-level capabilities such as zero-shot classification but reduced the cross-modal score to 43.4,
against 76.6--79.5 for the base, with the largest loss in the text space (R@10 19.3 against 75.7--81.2). With a detached target,
the own embeddings receive no gradient from the translation.

\paragraph{Contribution of the patch-level term.} For satellite images, removing the patch term reduced the patch translation score
from 19\,/\,29\% to approximately zero and the within-image structure from 0.89 to 0.65\,/\,0.79, while the class-token retrieval
and the probe remained within the margin. Mean-pooled translated patches lost most of their value as a substitute: the share of the
gap closed fell from 70 and 43\% to 36 and 18\% in the two directions, while the class-token substitution was unchanged. The patch
term thus provides the spatial component of the translation, at no measurable cost to the global component.

\paragraph{Temperature, batch size and statistics, and SIGReg weight.} A higher temperature ($T = 0.2$) favoured the patch
translation (44\,/\,50\%) at the expense of the class token in both applications, while a lower temperature ($T = 0.05$) had the
opposite effect and also reduced the share of the gap that an optical classifier recovers from translated radar images (37.3
against 65.6\%); $T = 0.1$ balances the two. Halving the batch size reduced retrieval for satellite images, where this run also doubled the SIGReg weight to keep the product of
weight and batch size constant, so that it changes two settings; for radiographs, where only the batch size changed, it
matched the base on the cross-modal score, yielded the highest zero-shot AUC and 5x200 accuracy among the single-change rows, and
required approximately 70\% of the training time, and it was therefore adopted there. Computing every batch-coupled term over the
global batch rather than per device makes the loss independent of the number of devices and changed the results within the seed
range. Halving the SIGReg weight instead of globalising the text invariance kept the cross-modal score within the margin but met
the alignment criterion (AUC 55.4, $p = 0.005$), and was not adopted. Several configurations classified as unaligned show a small
paired-cosine excess that the permutation test detects ($p \le 0.01$ for seed 44, the detached target, $T = 0.2$ and batch 128)
but that remains below the effect-size threshold of 55.

\section{Geometry of the Learned Translation}
\label{sec:rotation}

The hypothesis of Section~\ref{sec:background} is that the two own spaces differ mainly by their orientation, in which case a
translation between them can be close to an orthogonal map. Figure~\ref{fig:toy} shows, on synthetic data, that a simplified
skipless translator trained with an InfoNCE objective recovers a planted rotation (Appendix~\ref{app:toy}). We now examine whether
the class-token translators of the trained models are approximately rotations as well.

For each direction we consider the own class token of the source tower, the own class token of the target tower and the
translation produced by the source tower, normalised as in the loss, and fit three mappings from the own embedding of the source
to that of the target, evaluated out of sample: a scaled rotation (the orthogonal Procrustes solution with one scale factor,
centred, plus the target mean), the same map fitted on shuffled pairs as a chance control, and an unconstrained linear map (ridge
regression). The Procrustes solution is not constrained to a proper rotation and may include a reflection. Because the
parameter-free layer normalisation removes the all-ones direction, the fitted maps have a zero singular value, so that a reflection
can be turned into a proper rotation without changing the fit; constraining the determinant to $+1$ leaves every retrieval and $R^2$
value of Table~\ref{tab:rot} unchanged, and we therefore refer to the fitted maps as rotations. For radiographs and reports the
mappings are fitted on the 1,859 held-out pairs and evaluated by R@10 among the 202 validation pairs (image to report\,/\,report to
image, each in the target space; for the trained translator these are the text-space image-to-report and the image-space
report-to-image recall of Table~\ref{tab:vl-main}). For satellite images they are fitted on 8,000 BigEarthNet training pairs and evaluated by top-10 retrieval among the 4,000
test images of Section~\ref{sec:vv-results} (optical to radar\,/\,radar to optical).

\begin{table}[!t]
\centering\small
\caption{Out-of-sample retrieval and fit of the class-token translator against a scaled rotation, a shuffled-pair rotation
and a ridge map estimated from the source tower's own class token.}
\label{tab:rot}
\begin{adjustbox}{max width=\textwidth}
\begin{tabular}{lcc}
\toprule
 & chest radiograph $\leftrightarrow$ report & Sentinel-2 $\leftrightarrow$ Sentinel-1 \\
\midrule
\multicolumn{3}{l}{\emph{retrieval with the mapped source as query (R@10 / top-10, in percent)}} \\
untranslated own CLS & \rr{3.5}{4.0} & \rr{0.45}{0.25} \\
rotation fitted on shuffled pairs (chance) & \rr{3.0}{4.0} & \rr{0.30}{0.28} \\
\B{scaled rotation, own $\to$ own} & \B{\rr{77.2}{79.7}} & \B{\rr{82.8}{83.8}} \\
ridge map, own $\to$ own & \rr{74.8}{77.7} & \rr{83.0}{85.5} \\
\B{the trained translator} & \B{\rr{78.7}{78.7}} & \B{\rr{77.3}{79.2}} \\
\midrule
\multicolumn{3}{l}{\emph{the translator compared with rotations and linear maps}} \\
cos(translated, mapped own CLS) & \rr{0.852}{0.882} & \rr{0.886}{0.890} \\
\quad same, random pairs & \rr{0.028}{0.036} & \rr{0.142}{0.155} \\
$R^2$ of the translated CLS by a rotation fitted to it & \rr{0.925}{0.893} & \rr{0.968}{0.967} \\
$R^2$ of the translated CLS by a ridge map fitted to it & \rr{0.972}{0.973} & \rr{0.987}{0.989} \\
scale of the fitted rotation, own $\to$ own & \rr{0.60}{0.62} & \rr{0.782}{0.804} \\
\bottomrule
\end{tabular}
\end{adjustbox}
\end{table}

The upper block of Table~\ref{tab:rot} reports retrieval with the mapped source embedding as the query in the target space: without
any mapping, through a rotation fitted on shuffled pairs, through the rotation and the ridge map fitted between the two own
spaces, and through the trained translator. The lower block characterises the translator: the cosine between the translation of a
sample and its mapped own embedding, the same quantity for random pairs, the coefficient of determination $R^2$ with which an
rotation and a ridge map, each fitted from the own embedding of the source to its translation, reproduce the translation, and
the scale factor of the fitted rotation.

The results are consistent with the hypothesis in both applications, with two qualifications. First, the two own spaces are related
by an approximate rotation. A single scaled rotation of the entire space, estimated from the paired own embeddings
alone, retrieves the partner in 77.2\,/\,79.7\% of cases for radiographs and reports and in 82.8\,/\,83.8\% for satellite images,
whereas the untranslated embeddings and the map fitted on shuffled pairs remain at chance level. An unconstrained linear map performs
at most marginally better (74.8\,/\,77.7 and 83.0\,/\,85.5), so most of the linear relation between the two spaces is captured by an
rotation. Second, the translator is close to, but not identical with, a rotation. A linear map of the source explains 97
to 99\% of the variance of its output and a rotation fitted to it 89 to 97\%, leaving 3 to 11\% that a rotation does
not capture, and its output agrees with the rotation fitted between the own spaces to a cosine of 0.85 to 0.89, against 0.03
to 0.16 for random pairs. Since the translator is a nonlinear layer that is not constrained to be orthogonal, this is a property
of what it learned rather than of its design.

The two applications differ in one respect. For radiographs and reports, the trained translator retrieves about as well as the fitted
rotation (78.7\,/\,78.7 against 77.2\,/\,79.7). For satellite images, the rotation fitted after training on 8,000 pairs
retrieves 4 to 6 points better than the trained translator (82.8\,/\,83.8 against 77.3\,/\,79.2). The translator has thus converged
to a rotation-like map, but not to the best such map between the two spaces. Possible contributors, not tested here, are that the contrastive objective constrains the translation only as far as needed to separate the samples of a batch, whereas a
least-squares fit uses every pair in full; that the translator is trained on augmented crops and evaluated on full images; and that
the same translator layer must also produce the translated patch grid. Closing this gap during training is left to future work.

Bootstrap intervals over the evaluated queries (1,000 resamples) place the satellite gap outside sampling variability: 82.8 [81.7,
84.0] and 83.8 [82.7, 84.9] for the fitted rotation against 77.3 [75.9, 78.6] and 79.2 [78.0, 80.5] for the translator. For
radiographs and reports the intervals of the rotation and of the translator overlap, for instance 77.2 [71.3, 82.7] against
78.7 [72.8, 84.2].

To test whether independently trained encoders are already related by a rotation, we trained the two satellite encoders for
20k steps without the translation objective, as two independent LeJEPA encoders with a SIGReg weight of 0.05, and fitted the same
maps on the same fit and test images. For the jointly trained 20k model, a rotation between the own spaces retrieves the partner
in 70.4 and 69.5\% of cases (73.7 and 75.6\% after standardising every feature with the statistics of the fit set). For the
independently trained encoders, it retrieves the partner in 7.1 [6.4, 8.0] and 11.4 [10.4, 12.4]\% of cases (17.6 and 20.6\%
after standardisation), and a ridge map in 31.7 and 27.8\% (40.0 and 30.5\%). Independently trained encoders thus share some
structure, well above chance (0.25\%), but are related by a rotation far more weakly than the jointly trained ones: in this
comparison the near-rotational relation between the own spaces is produced by the joint training with the translation
objective rather than present in independently trained encoders.

Concurrent theoretical work arrives at a rotation from the opposite direction. \citet{balestriero2026platonic} prove that JEPAs
trained separately on two modalities learn the same representation up to one rotation when the positive pairs of each modality are
two observations of the same scene and the minimiser of the objective is unique up to rotation, which they establish for
second-order anti-collapse terms under a spectral gap and assume for SIGReg. When the positives are instead corruptions of a single
observation, as the crops of one image and the sentence subsets of one report from which our encoders learn, their result bounds the
misalignment without predicting a shared rotation, so the weak relation between our independently trained encoders does not
contradict it. Their theorem concerns encoders trained separately, whereas our observations concern encoders trained jointly; the
two results were obtained independently and point to the same form of relation. In our setting, the translation objective relates
two observations of the same scene, a radiograph and its report or the optical and radar images of one location, which the
crop-based views of either encoder alone do not, and with it the two own spaces became related by a rotation. The positive of each translation is the centroid $c^{m}_i$ of the global views of the
sample (Section~\ref{sec:cls}), which matches the translation to the observation as a whole rather than to one of its crops. The
centroid is still computed from a single observation, however, and averaging the crops of one observation cannot remove the
properties common to all of them, such as its viewpoint and lighting; the second observation of the scene therefore comes only from
the other modality. Whether encoders trained separately with two observations of each scene as positives,
for instance the frontal and lateral radiographs of one examination, would show this relation without the translation objective
remains open.

Two observations follow. First, the jointly trained encoders did not learn directly aligned spaces; within the precision of this analysis, they
learned similar neighbourhood structure in differently oriented frames, and the translator learned an approximately rotational change
of frame. Second, the translator type changed where this change of frame resided. With a residual translator the encoders became
aligned (Section~\ref{sec:ablation}), and their own spaces retrieved each other as well as a rotation fitted between them; with
a skipless translator the change of frame remained in the translator and the own spaces stayed unaligned. The fitted maps are a
diagnostic and not a component of the method; all results reported in this paper are obtained with the trained translator.

\section{Discussion}
\label{sec:discussion}

The results support three findings, each under the conditions tested here. First, cross-modal retrieval and classification can be
obtained without directly aligning the own embedding spaces of the two encoders: the paired-cosine test detects no alignment in the
final models, while the translation supports zero-shot classification, retrieval and the substitution of a missing sensor. Second,
the residual connections of the translator change how the model accommodates the cross-modal translation objective: with a residual
translator the encoders became strongly aligned, with a skipless one they did not in the final models and only weakly in one
ablation, at comparable cross-modal performance. Third, the learned
class-token translation is approximately a single rotation, a relation between the own spaces that arose with the joint objective and
was much weaker in a control trained without it, with a patch-level extension that transfers within-image similarity structure
for spatially corresponding images. None of these findings establishes a general rule for multimodal representation spaces.

These findings complement concurrent theoretical work by \citet{balestriero2026platonic}, developed independently of ours, which
establishes when JEPAs trained separately on different modalities learn the same representation up to a rotation. The two works
approach the geometry of multimodal representations from opposite directions, theirs from a theorem about separately trained
encoders and ours from experiments with a translator trained jointly with the encoders, and both arrive at a rotation as the form of
the relation between the spaces. DuoJEPA addresses a different question, namely how a cross-modal translation can be learned
end-to-end between own spaces that remain directly unaligned. In our setting, in which each encoder learns from crops of a single
observation, the near-rotational relation between the own spaces was much stronger with joint training than with separate training
(Section~\ref{sec:rotation}); although this rests on a single control, it offers an empirical perspective that complements their
analysis.

Retaining separate own spaces has practical implications. Each encoder is shaped by its own modality and by the requirement that the
other modality be reachable from it, so that its own embedding can serve single-modality tasks without a shared coordinate system.
The translation requires one additional layer at inference and is available for any input of either modality. The observed
trade-off for radiographs is a lower label efficiency of the image encoder relative to LeVLJEPA, in exchange for a substantial gain on
every cross-modal task.

\paragraph{Limitations.} Most ablation entries are single training runs; the variability across training seeds was measured only on
the base configuration of each application (three seeds for radiographs and reports, two for satellite images), and the final models
are single runs. Beyond bootstrap confidence intervals over the test samples, which do not capture training variability, no statistical testing
was performed for the comparisons between models. The alignment test uses 1,859 pairs from 645 patients, and an AUC near 50 shows only
the absence of detectable direct cosine alignment under this test. The radiograph-report model is trained for 20k steps on one dataset and evaluated on CheXpert-derived
benchmarks; the LeVLJEPA comparison relies on a reimplementation with deviations from the published recipe. The satellite models are
evaluated on a single benchmark, and our Copernicus-Bench evaluation differs from the official code in several respects
(Appendix~\ref{app:eval}), so that differences of a few tenths of a point with respect to published values are not interpretable. The
rotation analysis concerns the class-token translator and does not account for the patch-level translation, and its control without
the translation objective is a single run trained with a different SIGReg weight. The patch
metrics measure the transfer of representational structure, and neither pixel-level reconstruction nor the sensitivity of the patch
term to misregistration was evaluated. Translator designs other than the residual and skipless transformer layer (a linear map, an MLP
of similar size, frozen encoders, other initialisations) were not compared. Finally, both applications involve exactly two modalities.

\paragraph{Future directions.} Natural next steps are the translator comparisons listed above, repeated runs of the key ablations,
visual and quantitative analyses of patch correspondence and misregistration, and an extension to more than two modalities with one
translator per pair. A more speculative direction, which we have not tested, is the use of skipless translators in predictive models
of temporal dynamics, where the translation would operate between the patch representations of successive time steps.

\section{Conclusion}
\label{sec:conclusion}

DuoJEPA retains one self-supervised embedding space per modality and learns, within each encoder and jointly with both encoders, a
transformer layer that translates into the other space. In our experiments on radiographs with reports and on optical and radar
satellite images, the final models with a skipless translator kept the own spaces directly unaligned under a paired-cosine test while supporting
cross-modal retrieval, zero-shot classification and the substitution of a missing sensor, whereas a residual translator led the
encoders to align. The learned class-token translation is well approximated by a single scaled rotation, and for spatially
corresponding images a patch-level term transfers within-image similarity structure. Treating the relation between modalities as a
mapping to be learned, rather than a space to be shared, may be useful where each modality warrants a representation of its own; the
conditions under which this holds beyond the settings studied here remain to be established.

\section*{Data Availability}

CheXpert Plus and the CheXpert evaluation labels are available from the Stanford Center for Artificial Intelligence in Medicine and
Imaging under its research data use agreement \cite{chambon2024chexpertplus,irvin2019chexpert}; the data may not be redistributed.
SSL4EO-S12 \cite{wang2023ssl4eo} and BigEarthNet v2 \cite{clasen2024reben} are publicly available under their respective licences
(Copernicus Sentinel data, and the CDLA-Permissive-1.0 licence for BigEarthNet).

\section*{Acknowledgments}

This research was supported in part by the National Science Foundation under award number 2216140, which funded the Kentucky
Research Informatics Composable Cloud (KyRICC) computing infrastructure used in this work. Any opinions, findings, and
conclusions or recommendations expressed in this material are those of the authors and do not necessarily reflect the views of
the National Science Foundation.

\bibliographystyle{plainnat}
\bibliography{duojepa_paper}

\appendix
\section{Implementation Details}
\label{app:impl}

\begin{table}[!h]
\centering\small
\caption{Configuration of the two models.}
\label{tab:recipes}
\begin{adjustbox}{max width=\textwidth}
\begin{tabular}{lll}
\toprule
 & Sentinel-2 $\leftrightarrow$ Sentinel-1 & chest radiograph $\leftrightarrow$ report \\
\midrule
towers & two ViT-B/14, 4 registers, random init.: 12 bands / 2 bands & image ViT-B/14, 4 registers, random init.; text BERT-base \cite{gu2021pubmedbert}, pre-trained, 128 tokens \\
layers & 12 encoder + 1 skipless translator & 12 encoder + 1 skipless translator \\
translation & CLS InfoNCE + patch InfoNCE, $T$ 0.1 & CLS InfoNCE, $T$ 0.1 \\
global views & 2 per modality, 224 px, scale [0.4, 1], shared crop and flip & image 2 $\times$ 224 px, scale [0.5, 1], no flip; text 2 subsets, half to all sentences \\
local views & 4 per modality, 98 px, scale [0.05, 0.4] & image 4 $\times$ 98 px, [0.1, 0.5]; text 4 subsets of one or two sentences \\
batch, steps, seed & 256, 100k (warm-up 20k, ramp to 40k), seed 42 & 128, 20k (warm-up 5k, ramp to 7.5k), seed 42 \\
learning rates & towers $2\times10^{-4}$, translators $10^{-4}$, minimum $10^{-6}$ & image $2\times10^{-4}$, text $5\times10^{-5}$, translators $10^{-4}$, minimum $10^{-6}$ \\
LeJEPA & $\lambda$ 0.005 both, $M = 1024$ directions, projector $\to$ 512 & $\lambda$ 0.02 image / 0.005 text, $M = 2048$, projector $\to$ 1024 \\
hardware & 2 $\times$ H200, 28 h 49 min (node shared with other runs) & 2 $\times$ H200, 2 h 34 min \\
\bottomrule
\end{tabular}
\end{adjustbox}
\end{table}

All image towers are DINOv2-with-registers ViT-B/14 bodies \cite{oquab2024dinov2,darcet2024registers} with random initial weights;
the text tower is initialised from a biomedical BERT \cite{gu2021pubmedbert}. A 224-pixel view yields a $16\times16$ patch grid
(Table~\ref{tab:recipes}). The projector is an MLP with two hidden layers of width 2048, each followed by batch normalisation
synchronised across devices and a GELU activation, and receives the raw class token of $h_{12}$. SIGReg is evaluated with a 17-knot
trapezoidal rule on $s \in [-5, 5]$, with directions drawn afresh at every step (identically on all devices) and its empirical
characteristic function averaged across devices. Encoders and projectors follow a linear warm-up to their peak learning rate and a
cosine decay to $10^{-6}$; the learning rate of the translator layers is zero until the translation is activated at $W$, then rises
over $W$ steps to its peak at $2W$ and decays by a cosine schedule. We use AdamW with a weight decay of 0.04 on weight matrices,
clip gradient norms at 3.0 per tower, skip an update on all devices when any gradient is not finite, and train in bf16 with fully
sharded data parallelism. For satellite images, photometric perturbations are drawn independently per sensor: for the optical tower
a brightness and contrast jitter (probability 0.8), a band-mean grey-level conversion (probability 0.2) and Gaussian blur
(probability 0.5); for the radar tower only the last two. Satellite inputs are standardised per band with the pre-training
statistics; radiographs are converted to three channels and normalised with the ImageNet statistics. Four of the ten satellite
ablation runs were trained on H100 GPUs with gradient checkpointing, the others on H200 GPUs.

\paragraph{Skipless initialisation.} Following \citet{ji2025skipless}, with $U, V$ orthogonal matrices from the singular value
decomposition of a Gaussian matrix we set $W_V = cU$ and $W_O = cV^{\top}$, and $W_Q W_K^{\top} = \alpha Z + \beta I$ with
$Z_{kl} \sim \mathcal{N}(0, 1/d)$, factorised by its singular value decomposition; the MLP weights are orthogonal, the biases zero,
and the normalisation layers have unit gain and zero bias, with $\alpha = 2$, $\beta = 0.6$, $c = 3$ and a different seed per tower;
the same initialisation is used for the text tower. The skipless forward pass modifies the computation rather than the weights, so
that a checkpoint records it in its configuration and every reader reconstructs it before reading the translated embedding; the own
embedding is unaffected. At inference each tower is evaluated through its standard forward pass, with the own embedding read at layer
12 and the translated embedding at layer 13.

\section{Evaluation Protocols}
\label{app:eval}

\paragraph{Chest radiographs and reports.} Test images are decoded as grey-level images, resized to 224 pixels on the short side
(bicubic) and centre-cropped; texts are tokenised and padded or truncated to 128 tokens. Zero-shot classification follows CheXzero
\cite{tiu2022chexzero}: for each of the five findings, the prompts ``$\langle$finding$\rangle$'' and ``no $\langle$finding$\rangle$''
are encoded by the text tower and the two cosines are converted into a probability by a softmax. In the image space, the translated
embeddings of the prompts are compared with the own embedding of the image; in the text space, the translated embedding of the image
is compared with the own embeddings of the prompts. For 5x200 we use the prompts of GLoRIA \cite{huang2021gloria} (five per class),
take the maximum similarity over the prompts of a class, standardise per class and predict the class with the largest score; for
GLoRIA itself only the global similarity is used (its published 5x200 accuracy additionally uses local similarities). The linear
probe is trained on the frozen own embedding of the image tower with binary cross-entropy (Adam, learning rate $10^{-4}$, batch 64,
at most 50 epochs with early stopping on 5,000 held-out training images) on the pre-training and hold-out images with their CheXpert
labels, and tested on the 202 validation images. The alignment test is described in Section~\ref{sec:eval}.

\paragraph{Optical and radar satellite images.} The BigEarthNet images of 120 pixels are resized to 224 pixels (bilinear) and
standardised with the pre-training statistics. The probe follows Copernicus-Bench \cite{wang2025copernicusfm}: 11,894\,/\,6,117\,/\,5,991
training, validation and test images, a linear classifier on the frozen embedding, SGD with a batch of 64 for 50 epochs, the learning
rate selected on the validation set from $\{10^{-4}, 10^{-3}, 10^{-2}, 10^{-1}, 1, 10\}$, three runs, and micro mAP. Our implementation
differs from the official code in five respects: the subsample is drawn at random for each run (seeds 0, 1, 2) instead of being read
from the official 5\% split file; the optimiser uses momentum 0.9 without weight decay and with a constant learning rate, instead of
no momentum, weight decay $10^{-4}$ and a one-cycle cosine schedule; the micro AP is computed once over the entire test set rather
than as a batch-weighted mean of per-batch values; the input resizing and normalisation are our own; and we read the class token
where each published model uses its own global readout. In addition, the classifier initialisation and batch order are drawn from a
shared random stream that is not reseeded per run, which adds a variability of about $\pm$0.5 mAP beyond the variation over the three
subsamples; comparisons between our models use identical draws. These differences do not affect comparisons among our own models,
which share one protocol. The retrieval, patch translation and within-image structure scores are computed on 4,000 clean test images
drawn with a fixed seed; the rotations of Section~\ref{sec:rotation} are fitted on 8,000 clean training images drawn with
another fixed seed.

\section{Confidence Intervals}
\label{app:ci}

\begin{table}[!h]
\centering\small
\caption{Point estimates and 95\% bootstrap confidence intervals of the radiograph--report results over the test samples, with the
paired differences between DuoJEPA and the reimplemented LeVLJEPA.}
\label{tab:ci}
\begin{adjustbox}{max width=\textwidth}
\begin{tabular}{lccccc}
\toprule
metric & DuoJEPA & LeVLJEPA (reimpl.) & CheXzero & GLoRIA & DuoJEPA $-$ LeVLJEPA \\
\midrule
zero-shot AUC, image space & 86.6 [84.8, 88.3] & 60.0 [56.5, 63.6] & 88.2 [86.5, 89.9] & 64.7 [61.9, 67.4] & $+26.6$ [22.8, 30.4] \\
zero-shot AUC, text space & 86.3 [84.4, 88.1] & 62.6 [58.7, 66.3] & -- & -- & $+23.7$ [19.8, 27.9] \\
5x200 accuracy & 47.5 [44.4, 50.8] & 39.9 [36.7, 43.0] & 46.3 [43.3, 49.5] & 58.8 [55.6, 61.8] & $+7.6$ [3.9, 11.3] \\
5x200 P@5 & 56.0 [53.7, 58.2] & 49.8 [47.6, 51.9] & 48.3 [45.9, 50.7] & 60.9 [58.5, 63.2] & $+6.2$ [4.5, 8.0] \\
R@10 i$\to$t, image space & 78.2 [72.3, 83.7] & 63.4 [56.4, 69.8] & 41.1 [34.2, 47.5] & 86.1 [81.2, 90.6] & $+14.9$ [7.9, 21.8] \\
R@10 t$\to$i, image space & 78.7 [72.8, 84.2] & 62.4 [55.9, 68.8] & 42.1 [35.2, 49.0] & 84.7 [79.7, 89.6] & $+16.3$ [9.4, 22.8] \\
R@10, text space & 78.7 [72.8, 84.2] & 66.8 [60.4, 73.3] & -- & -- & $+11.9$ [5.0, 18.8] \\
cross-modal score & 78.5 [73.1, 83.5] & 64.2 [58.4, 70.0] & -- & -- & $+14.4$ [8.6, 20.0] \\
alignment AUC & 48.3 [46.1, 50.3] & 51.0 [49.3, 52.8] & 88.7 [87.7, 89.7] & 98.9 [98.6, 99.1] & $-2.8$ [$-5.6$, 0.1] \\
\bottomrule
\end{tabular}
\end{adjustbox}
\end{table}

The intervals of Table~\ref{tab:ci} are percentile intervals from 2,000 bootstrap resamples (fixed seed) with each sample's prediction
held fixed: the 500 test images for zero-shot classification, the 1,000 images for 5x200, the 202 validation pairs as queries
against the fixed gallery for retrieval (the cross-modal score resampling its three recalls jointly by pair), and the 645 hold-out
patients for the alignment AUC. The differences use the same resamples for both models. The intervals describe the variability of
the evaluation samples only; each model is a single training run. For the satellite models, the intervals of the retrieval values
of Section~\ref{sec:rotation} are obtained in the same way over the 4,000 test queries (1,000 resamples).

\section{Synthetic Rotation Experiment}
\label{app:toy}

The experiment of Figure~\ref{fig:toy} uses three-dimensional unit vectors grouped into 40 clusters whose centres lie on a jittered
Fibonacci lattice on the sphere (jitter 0.15), with an angular spread of 4 degrees per cluster and cluster sizes drawn from a
Dirichlet distribution with parameter 2. Space B is obtained from space A by a rotation of 70 degrees about the axis $(1, 1, 0.3)$
(Rodrigues' formula). Thirteen clusters, chosen by a random permutation (seed 0), are held out; 16,384 training samples are drawn from
the remaining 27. The translator is a simplified single-token version of the skipless layer: two $3 \times 3$ linear maps (value and
output) followed by an MLP $3 \to 12 \to 3$ with a GELU activation, without normalisation layers or scale vectors and with the default
PyTorch initialisation; it is not constrained to be a rotation. It is trained with an InfoNCE objective over the batch (temperature
0.1, batch 256, one view, no layer normalisation) between the normalised translations of space-A samples and the normalised
space-B samples, using AdamW (learning rate $3\times10^{-3}$, no weight decay, cosine schedule) for 3,000 steps. Evaluated with an
orthogonal Procrustes fit on 4,096 fresh samples from the seen clusters, the learned map has a rotation angle of 70.05 degrees (true
value 70.00) and an axis error of 0.42 degrees; the cosine between the translator output and the true rotation is 0.992 on seen and
0.993 on unseen clusters (0.057 before training), and a linear fit to the translator has determinant $+0.993$, that is, a proper
rotation. The experiment plants the rotation and serves as an illustration, not as evidence about the trained models.

\end{document}

%% file: duojepa_patch_nce_figure_generic.tex
\begin{tikzpicture}[x=1cm,y=1cm,font=\small]
  \definecolor{he}{HTML}{5B3E96}\definecolor{hesoft}{HTML}{F2F0F7}
  \definecolor{tau}{HTML}{A0601A}
  \definecolor{tr}{HTML}{2E7D7A}\definecolor{trsoft}{HTML}{EAF2F2}
  \definecolor{ink}{HTML}{2B2836}\definecolor{muted}{HTML}{5C5969}
  \def\s{0.32}           
  \def\n{8}              
  \def\ri{3}\def\ci{4}   
  \pgfmathsetmacro{\side}{\n*\s}
  \pgfmathsetmacro{\yk}{(\n-\ri-0.5)*\s}   
  \pgfmathsetmacro{\xk}{(\ci+0.5)*\s}      
  \fill[trsoft] (0,0) rectangle (\side,\side);
  \draw[tr!40,line width=0.3pt,step=\s] (0,0) grid (\side,\side);
  \draw[tau,line width=1.2pt] (0,0) rectangle (\side,\side);
  \fill[tr] (\ci*\s,{(\n-\ri-1)*\s}) rectangle ({(\ci+1)*\s},{(\n-\ri)*\s});
  \node[white,font=\scriptsize\bfseries] at (\xk,\yk) {$k$};
  \node[align=center,anchor=south] at (\side/2,\side+0.5) {\textbf{\textcolor{tau}{source $\bar m$}, translated}\\[-1pt]
    {\color{muted}\footnotesize image $r$, layer 13, in the space of \textcolor{he}{$m$}}\\[-1pt]
    {\color{muted}\footnotesize anchors $t_{r,k}$, gradient flows}};
  \def\ox{5.9}
  \begin{scope}[shift={(\ox,0)}]
    \fill[hesoft] (0,0) rectangle (\side,\side);
    \draw[he!40,line width=0.3pt,step=\s] (0,0) grid (\side,\side);
    \draw[he,line width=1.2pt] (0,0) rectangle (\side,\side);
    \fill[he] (\ci*\s,{(\n-\ri-1)*\s}) rectangle ({(\ci+1)*\s},{(\n-\ri)*\s});
    \node[white,font=\scriptsize\bfseries] at (\xk,\yk) {$k$};
    \node[align=center,anchor=south] at (\side/2,\side+0.5) {\textbf{\textcolor{he}{target $m$}, own, same image}\\[-1pt]
      {\color{muted}\footnotesize image $r$, layer 12, detached}\\[-1pt]
      {\color{muted}\footnotesize positive $o_{r,k}$}};
  \end{scope}
  \def\nx{11.8}
  \foreach \d in {2,1,0} {
    \begin{scope}[shift={({\nx+0.22*\d},{0.22*\d})}]
      \fill[white] (0,0) rectangle (\side,\side);
      \fill[hesoft] (0,0) rectangle (\side,\side);
      \draw[he!40,line width=0.3pt,step=\s] (0,0) grid (\side,\side);
      \draw[he!70,line width=0.9pt] (0,0) rectangle (\side,\side);
      \fill[muted!55] (\ci*\s,{(\n-\ri-1)*\s}) rectangle ({(\ci+1)*\s},{(\n-\ri)*\s});
    \end{scope}
  }
  \node[white,font=\scriptsize\bfseries] at (\nx+\xk,\yk) {$k$};
  \node[align=center,anchor=south] at ({\nx+\side/2+0.22},{\side+0.5}) {\textbf{\textcolor{he}{target $m$}, own, other images}\\[-1pt]
    {\color{muted}\footnotesize other samples of the batch, detached}\\[-1pt]
    {\color{muted}\footnotesize negatives $o_{r',k}$, same position $k$}};
  \draw[tr,line width=1.3pt,-latex] ({(\ci+1)*\s+0.04},\yk) -- ({\ox+\ci*\s-0.04},\yk);
  \node[anchor=south,font=\footnotesize] at ({(\side+\ox)/2},{\yk+0.05}) {pull together};
  \draw[tr,line width=1.0pt,-latex,dashed] (\xk,{(\n-\ri-1)*\s-0.02}) -- (\xk,-0.45) -- ({\nx+\xk},-0.45) --
    ({\nx+\xk},{(\n-\ri-1)*\s-0.02});
  \node[anchor=north,font=\footnotesize] at ({(\xk+\nx+\xk)/2},-0.5) {push apart};
  \node[draw=ink,line width=0.9pt,rounded corners=3pt,fill=white,align=center,inner sep=5pt,anchor=north]
    at ({(\nx+\side)/2+0.2},-1.05) {$\displaystyle \mathcal{L}^{m}_{\mathrm{patch}}
      = -\frac{1}{BNP}\sum_{r}\sum_{k=1}^{P}\log
      \frac{\exp\!\big(\langle t_{r,k}, o_{r,k}\rangle / T\big)}{\sum_{r' \in \mathcal{R}(r)} \exp\!\big(\langle t_{r,k}, o_{r',k}\rangle / T\big)}$\\[4pt]
      {\color{muted}\footnotesize $\mathcal{R}(r)$: image $r$ and every image of the other samples (the same sample's other view left out).}\\[1pt]{\color{muted}\footnotesize
      Mirror term in the space of \textcolor{tau}{$\bar m$}: roles swapped.}};
\end{tikzpicture}